\pdfoutput=1

\documentclass[11pt]{article}

\usepackage[final]{acl}

\usepackage{times}
\usepackage{latexsym}

\usepackage[T1]{fontenc}

\usepackage[utf8]{inputenc}

\usepackage{microtype}

\usepackage{inconsolata}

\usepackage{graphicx}

\usepackage{amsmath}
\usepackage{amssymb}
\usepackage{amsthm}

\usepackage{multirow}
\usepackage{booktabs}

\usepackage{pifont}

\usepackage{stfloats}
\usepackage{float}
\usepackage{caption}
\usepackage{soul}

\usepackage{pdflscape}

\usepackage{subfigure}
\usepackage{hyperref}
\usepackage{tabularx}
\usepackage{xspace}
\title{HEAL: A Hypothesis-Based Preference-Aware Analysis Framework}

\author{
 \textbf{Yifu Huo\textsuperscript{1}},
 \textbf{Chenglong Wang\textsuperscript{1}},
 \textbf{Qiren Zhu\textsuperscript{1}},
 \textbf{Shunjie Xing\textsuperscript{1}},
 \textbf{Tong Xiao\textsuperscript{1,2}\thanks{\xspace Corresponding author.}},
\\
 \textbf{Chunliang Zhang\textsuperscript{1,2}},
 \textbf{Tongran Liu\textsuperscript{3}},
 \textbf{Jingbo Zhu\textsuperscript{1,2}},
\\
 \textsuperscript{1}School of Computer Science and Engineering, Northeastern University, Shenyang, China \\
 \textsuperscript{2}NiuTrans Research, Shenyang, China \\
 \textsuperscript{3}CAS Key Laboratory of Behavioral Science, Institute of Psychology, CAS, Beijing, China
\\
\ttfamily{\{ifnoct, clwang1119\}@gmail.com}, 
\ttfamily{\{xiaotong, zhujingbo\}@mail.neu.edu.cn}
}

\begin{document}
\maketitle
\begin{abstract}

Preference optimization methods like DPO have achieved remarkable performance in LLM alignment.
However, the evaluation for these methods relies on a single response and overlooks other potential outputs, which could also be generated in real-world applications within this hypothetical space.
To address this issue, this paper presents a \textbf{H}ypothesis-based Pr\textbf{E}ference-aware \textbf{A}na\textbf{L}ysis Framework (HEAL), a novel evaluation paradigm that formulates preference alignment as a re-ranking process within hypothesis spaces. 
The framework incorporates two complementary metrics: ranking accuracy for evaluating ordinal consistency and preference strength correlation for assessing continuous alignment.
To facilitate this framework, we develop UniHypoBench, a unified hypothesis benchmark constructed from diverse instruction-response pairs. Through extensive experiments based on HEAL, with a particular focus on the intrinsic mechanisms of preference learning, we demonstrate that current preference learning methods can effectively capture preferences provided by proxy models while simultaneously suppressing negative samples.
These findings contribute to preference learning research through two significant avenues. Theoretically, we introduce hypothesis space analysis as an innovative paradigm for understanding preference alignment. Practically, HEAL offers researchers robust diagnostic tools for refining preference optimization methods, while our empirical results identify promising directions for developing more advanced alignment algorithms capable of comprehensive preference capture. Code and data available at \href{https://github.com/if-noc/HEAL}{this link}.

\end{abstract}

\section{Introduction}


Direct preference optimization (DPO) has emerged as the predominant method for aligning large language models (LLMs) with human preferences \cite{rafailov2024direct,xiao2025foundations}. 
Recent research on DPO has also explored various variants, including SimPO \cite{Meng2024SimPOSP}, ORPO \cite{Hong2024ORPOMP}, and KTO \cite{Ethayarajh2024KTOMA}.
 To evaluate the effectiveness of these preference alignment methods, researchers typically rely on benchmarks such as AlpacaEval \cite{Dubois2024LengthControlledAA} and MT-Bench \cite{Bai2024MTBench101AF}.
In the evaluation process using these benchmarks, we typically follow a standard procedure: given a prompt, we first generate a response from the aligned model using a temperature-based sampling method  \cite{Gu2024ASO}. 
Next, we employ a proxy model (such as GPT-4) to compare this response with the reference response, evaluating whether the model is effectively aligned.

However, this procedure faces a fundamental limitation because this \textit{sampling-based} evaluation approach only examines single responses sampled from target LLMs.
But in practice, the LLMs are commonly expected to generate a wide spectrum of diverse responses, which cannot be sufficiently covered by several sampled responses.
This misalignment between evaluation and real-world LLM development prohibits researchers and developers from assessing the LLMs' performance accurately.
Furthermore, this paradigm also neglects the relative comparison of responses, which is fundamentally modeled in preference learning scenarios.

To address these limitations in the evaluation, we propose HEAL (\underline{\textbf{H}}ypothesis-based pr\underline{\textbf{E}}ference-aware \underline{\textbf{A}}na\underline{\textbf{L}}ysis), a novel framework that evaluates LLMs through the lens of ranked hypothesis spaces.
Inspired by the ranking-based evaluation approaches such as RewardBench \cite{lambert2024rewardbench}, HEAL conceptualized preference alignment as a dynamic re-ranking process within the hypothesis space, enabling comprehensive assessment through two complementary quantitative methods: 
(1) The first metric is ranking accuracy, which is measured via Kendall's Tau between the ranking of policy model and proxy preference model (used for training data annotation).
This metric evaluates ordinal consistency in preference learning and directly assesses whether the model preserves the relative ordering of hypotheses as intended by the preference signals.
(2) The second one is preference strength correlation.
We use Pearson correlation between generation likelihoods and gold-standard preference scores as the metric.
This metric captures continuous alignment precision.
Unlike binary or ordinal measures, it quantifies the model's sensitivity to fine-grained preference distinctions, ensuring quantified relationships in preference strength are maintained across hypotheses.

We evaluate mainstream preference learning methods using HEAL to address three key research questions.
(RQ1): \textit{Do these methods effectively capture preference information?}
While ranking accuracy confirms that LLMs acquire preferences through optimization, results reveal incomplete learning.
To elaborate, current methods struggle to absorb all preference signals fully.
(RQ2): \textit{Can LLMs discern proxy model-specific preferences?}
Experiments demonstrate that LLMs successfully learn distinct preference patterns from different proxy models, showcasing HEAL's sensitivity to subtle inter-model variations.
(RQ3): \textit{How do learned preferences vary across methods?} All tested methods achieve strong in-distribution alignment with proxy models, but out-of-distribution performance degrades significantly, except for SimPO, which exhibits notable generalization. 
These results validate the partial efficacy of preference alignment while underscoring critical limitations, particularly in robustness and completeness of learned preferences.

Our main contributions are:
\begin{itemize}
    \item To the best of our knowledge, we are the first to present a systematic study assessing how effectively LLMs capture proxy model preferences through a hypothesis lens.
    \item We construct HEAL, a hypothesis-based preference-aware analysis framework that quantifies the preference modeling analysis into two metrics: ranking accuracy and preference strength correlation.
    Furthermore, we construct a \underline{\textbf{Uni}}fied \underline{\textbf{Hypo}}thesis \underline{\textbf{Bench}}mark (UniHypoBench) to support the evaluation pipeline of HEAL.
    \item We conduct comprehensive experiments using HEAL.
    The experimental results yield three key observations:
    (1) Low ranking accuracies reveal a significant preference learning gap, which we attribute to the task's substantially greater complexity compared to standard pairwise response ranking;
    (2)  Current LLMs struggle to capture nuanced preference strength relationships between hypotheses, a limitation evidenced by low correlation scores on our novel preference strength metric;
    (3) We uncover distinct, model-specific preference signatures, demonstrating that each model learns and prioritizes values differently.
\end{itemize}

\section{Preliminary}
\subsection{Sequence Likelihood}
\label{sec:sequence-likelihood}
In the literature of LLMs, a model parameterized with $\theta$ is essentially a generative large language model, which is applied to generate a response sequence $y$ when prompted with input $x$.
The response $y$ is typically generated by sampling the next tokens auto-regressively from a probabilistic distribution.
At each time step in this procedure, the model selects the next token randomly to form a new input.
Under this approach, the likelihood of generating a specific sentence can be obtained by computing the conditional probability, which can be written as:
{
\begin{eqnarray}
    \pi_\theta(y | x) = \prod^{|y|}_{n=0}P_\theta(y_n|y_{<n},x)
\end{eqnarray}
}
\noindent where the term $P_\theta(y_n|y_{<n},x)$ represents the probability of the $n\mathrm{-th}$ token of response $y$.
The sequence likelihood reflects how an LLM tends to generate a specific response, and this likelihood also serves as a core component of other metrics, such as perplexity (PPL).

\subsection{Human Preference Alignment}
\label{sec:alignment-of-llms}
In the realm of LLMs, the training process generally encompasses three key stages: pre-training, supervised fine-tuning, and human preference alignment \cite{Ouyang2022TrainingLM,bai2022training,wang2024hybrid}.
Recent advancements have demonstrated that alignment can be effectively achieved through two branches of RLHF: reward-based methods and reward-free methods.
Throughout the training process, LLMs inherently learn human preferences either through reward scores provided by a reward model \cite{wang2025gram,wang2025rovrm} or by utilizing ranked pairs of responses \cite{rafailov2024direct,Meng2024SimPOSP}.
However, the reward-based methods often require extensive reward modeling and face challenges in scalability and generalization \cite{Gao2022ScalingLF}.
Consequently, recent work has shifted towards reward-free methods, which directly optimize preferences without explicit reward signals.
This shift highlights the growing importance of reward-free approaches in addressing the limitations of traditional reward-based methods, offering a more scalable and efficient path for aligning LLMs with human preferences.

\paragraph{Reward Modeling.}
To effectively capture human preferences, a widely adopted approach involves training a reward model using human preference datasets.
In the context of RLHF, a reward model is generally formulated as a function $r_\phi(x, y)$, where $\phi$ represents model parameters, $x$ denotes the instruction, and $y$ corresponds to the response. 
To develop such a reward model, a foundation LLM is optimized by minimizing the Bradley-Terry loss \cite{Stiennon2020LearningTS}, as follows:
\begin{multline}
    \mathcal{L}_\mathrm{reward} = -\mathbb{E}_{(x,y_{w},y_l)\sim  \mathcal{D}_\mathrm{p}}\log( 
    \\  \sigma(r_{\phi}(x,y_{w})-r_{\phi}(x,y_{l}))) \label{eq:reward-training}
\end{multline}
\noindent Here, $D_\mathrm{p}$ represents the human preference dataset, which comprises input tuples containing an instruction $x$ and a pair of responses $(y_w,y_l)$ with preference $y_w \succ y_l$, where $y_w \succ y_l$ indicates that $y_w$ is preferred over $y_l$ according to human or model-based annotations.
This dataset serves as the foundation for training the reward model, enabling it to capture human preferences effectively.

Although it has been discussed that recent work has increasingly focused on reward-free methods, reward models continue to play a significant role in alignment.
For instance, a robust reward model can act as a reliable human proxy, which is capable of constructing high-quality preference data for reward-free methods such as DPO.
The training of reward models lays the groundwork for understanding and optimizing human preferences, which will be further explored in the context of preference optimization in subsequent sections.

\paragraph{Preference Optimization.}
Building on the discussion of reward models and their role in alignment, Rafailov et al. \cite{rafailov2024direct} introduced DPO, a novel approach that inherently integrates the reward model within the policy model itself.
In DPO, the policy model is directly optimized using a preference dataset, which can also serve as the basis for training a reward model.
This dual-purpose utilization of the dataset highlights the flexibility and efficiency of the DPO. The DPO loss function can be given by:
\begin{align}
    &\mathcal{L}_\text{DPO}(\pi_{\theta}) = -\mathbb{E}_{(x, y_w, y_l)\sim \mathcal{D}_\mathrm{p}} \nonumber
    \\ &\left[\log \sigma \left(\beta \log \frac{\pi_{\theta}(y_w | x)}{\pi_\mathrm{ref}(y_w | x)} - \beta \log \frac{\pi_{\theta}(y_l | x)}{\pi_\mathrm{ref}(y_l | x)}\right)\right]\label{eq:dpo-loss}
\end{align}
\noindent where $\beta$ denotes a parameter that controls the strength of constraints, ensuring the optimized policy $\pi_\theta(y|x)$ does not deviate excessively from the reference policy $\pi_{ref}(y|x)$.
Notably, there is a significant conceptual similarity between the loss functions of DPO and reward modeling loss as introduced in Eq. \ref{eq:reward-training}.
Although the specific objectives of Eq. \ref{eq:reward-training} and Eq. \ref{eq:dpo-loss} differ in formulation, their underlying goals are fundamentally aligned: \textit{Maximize the likelihood of generating responses preferred by humans while minimizing the probability of producing dispreferred ones}.
In conclusion, understanding the shared principles of preference modeling between reward-based and reward-free methods is crucial for uncovering the fundamental mechanisms of alignment in LLMs. 

\begin{figure*}[t]
    \centering
    \includegraphics[width=\linewidth]{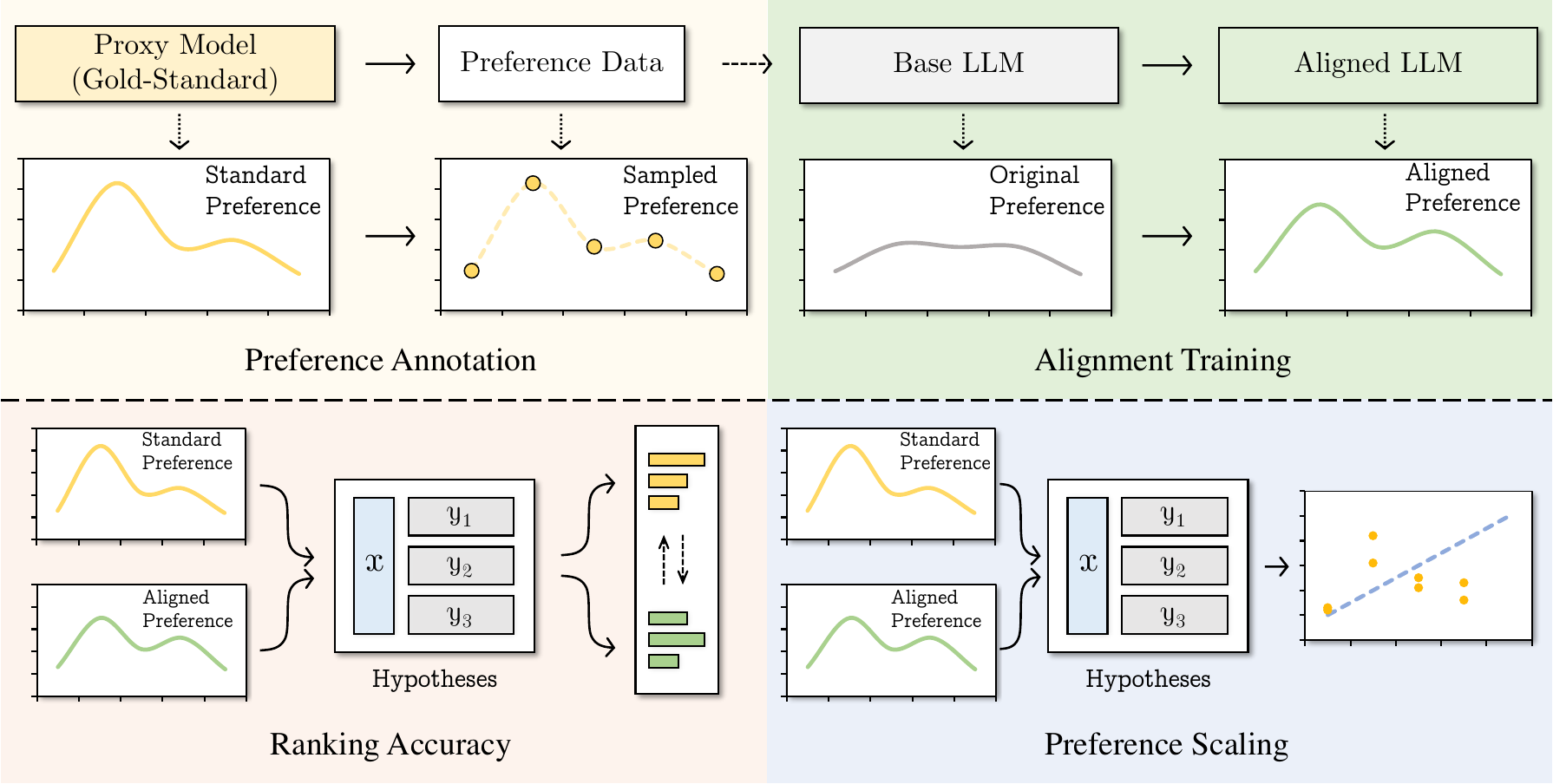}
    \caption{The overview of HEAL.
    We conceptualize preference learning as an alignment process between the original preference pattern and the standard preference sampled from the proxy model.
    The framework employs two evaluation metrics - ranking accuracy and preference strength correlation.    
    }
    \vspace{-0.3cm}
    \label{fig:main-figure}
\end{figure*}

\section{HEAL: A Hypothesis-based Analysis Framework}

We propose HEAL (hypothesis-based analysis), illustrated in Figure \ref{fig:main-figure}.
The framework models preference patterns as ranked hypothesis spaces and evaluates them through two complementary metrics:
(1) ranking accuracy for ordinal consistency and
(2) preference strength correlation for continuous alignment, which we detail in this section respectively.

\subsection{Ranking in Hypothesis Space}
\label{sec:alignment-reranking}
In the current evaluation procedure, recent research studies the behavior of LLMs directly from their generated content.
However, as mentioned in Sec. \ref{sec:sequence-likelihood}, generation is naturally a random process that hardly produces stable outputs.

\theoremstyle{definition}
\newtheorem{definition}{Definition}
\begin{definition}[\textit{Hypothesis Space}]

In the generation phase, the responses possibly differ within a constrained set $Y_x$ due to variations in hyperparameter configurations.
To study these responses, we extend the term hypothesis to LLMs by analogy to natural language understanding (NLU), where this terminology denotes a candidate sentence sampled from the output space \cite{Proebsting2024HypothesisonlyBI}.
Here, we consider the constrained set $Y_x$ as the hypothesis space for the given $x$, which contains all possible responses (\textit{i.e. }hypotheses). 
The hypothesis space is formulated as follows:
{
\begin{multline}
    Y_x = \{ y_i \in \bigcup^{\infty}_{n=0}V^n \thinspace | \thinspace \mathbf{I}(x,y_1) \geq \cdots
    \\ \geq \mathbf{I}(x,y_i) \geq \cdots, \thinspace i \in N_+  \}
\end{multline}
}

\end{definition}

\noindent where $V$ denotes the vocabulary set defining the hypothesis space.
Formally, $Y_x$ constitutes a set comprising (potentially infinite) textual hypotheses generated from $V$.
These hypotheses are ordered by an indicator function $\mathbf{I}(x,y)$, which assigns a comparable scalar value to each hypothesis $y$.
Typical instantiations of $\mathbf{I}(x,y)$ include the generation probability $\pi_\theta(y|x)$ under the model parameter $\theta$, or a preference score given by a human annotator (or a strong LLM).
With these instantiations, the hypothesis space $Y_x$ is structured as an ordered set, where hypotheses are ranked in descending order.
Consequently, hypotheses positioned earlier in $Y_x$ exhibit a higher likelihood of being selected during generation.

\begin{definition}[\textit{Gold-Standard Hypothesis Space}]

From the perspective of generation probabilities, the alignment algorithms optimize the generation probabilities to favor preferred responses while suppressing the dispreferred ones, thus we can conclude alignment as reordering the hypothesis space $Y_x$ to better match a gold-standard hypothesis space $Y_{\mathrm{x;gold}}$.
Formally, the gold-standard hypothesis space $Y_{\mathrm{x;gold}}$ is defined as:
\begin{multline}    
    Y_{\mathrm{x;gold}}=\{ y_i \in \bigcup^{\infty}_{n=0}V^n \thinspace | \thinspace \mathrm{GS}(x,y_1) \geq \cdots 
    \\ \geq \mathrm{GS}(x,y_i) \geq \cdots, \thinspace i \in N_+  \}
\end{multline}

\noindent where the gold scoring function $\mathrm{GS}(x,y)$ quantifies the alignment quality of response $y$ to instruction $x$, as evaluated by either reward models or human annotators \cite{lambert2024rewardbench,Zhou2023LIMALI}.
From a formal perspective, $\mathrm{GS}(x,y)$ represents a specialized instantiation of the indicator function $\mathbf{I}(x,y)$, optimized to reflect ideal human preferences.
In conclusion, this space serves as the theoretical optimum for alignment objectives.

\end{definition}

\subsection{Quantitative Analysis Method}

\paragraph{Ranking Accuracy.}
In Section \ref{sec:alignment-reranking}, we formally defined both the hypothesis space $Y_x$ and the gold-standard hypothesis space $Y_{\mathrm{x;gold}}$. These spaces differ only in their internal ranking criteria, while their elements remain identical \cite{Chen2024PreferenceLA}.

To quantify the alignment quality between an LLM's outputs and gold-standard preferences, we propose measuring the ordinal discrepancy between $Y_x$ and $Y_{\mathrm{x;gold}}$.
Specifically, we adapt Kendall's Tau-b correlation coefficient as our metric for comparing their partial orders.
The Kendall's Tau-b statistic is formally expressed as:
\vspace{-0.1cm}
{
\begin{multline}
    \tau_b(Y_{x}^{(1)},Y_{x}^{(2)})
     = \\ \frac{\mathrm{C}(Y_{x}^{(1)},Y_{x}^{(2)})-\mathrm{D}(Y_{x}^{(1)},Y_{x}^{(2)})}{\sqrt{(\mathrm{T}_0-\mathrm{T}_1(Y_{x}^{(1)}))(\mathrm{T}_0-\mathrm{T}_2(Y_{x}^{(2)}))}}\label{eq:kendall-tau}
\end{multline}
}

\noindent where:
\begin{itemize}
    \item $Y_{x}^{(1)}$ and $Y_{x}^{(2)}$ denote two hypothesis spaces sharing identical response elements but potentially differently ordered by their respective indicator functions $\mathbf{I}_1(x,y)$ and $\mathbf{I}_2(x,y)$.
    \item $\mathrm{C}(\cdot,\cdot)$ counts concordant pairs - cases where the relative ordering of $(y_i,y_j)$ is consistent between both spaces, while $\mathrm{D}(\cdot,\cdot)$ counts discordant pairs with contradictory orderings.
    \item $\mathrm{T}_0=\binom{n}{2}$ represents the total possible pairs. 
    \item $\mathrm{T}_1(\cdot)$ and $\mathrm{T}_2(\cdot)$ are tie correction terms for each hypothesis space.    
\end{itemize}

The ranking of responses $y_i$, $y_j$ in each space is determined by comparing their indicator values $\mathbf{I}(x,y_i)$ and $\mathbf{I}(x,y_j)$.
The denominator's adjustment for ties ensures robustness when the indicator function produces discrete scores.
This metric provides a unified comparison capability, applicable to any system generating comparable $\mathbf{I}(x,y_i)$ values.

\begin{definition}[\textit{Ranking Accuracy}]
Furthermore, we notice that Kendall's Tau differs from the ranking accuracy in range.
Therefore, we map the original metric to obtain an accuracy ratio, as follows:
\begin{multline}
    \mathrm{RA}(\mathcal{D})
     = \\ \mathbb{E}_{(x,Y_{x}^{(1)},Y_{x}^{(2)}) \sim \mathcal{D}} \frac{\tau_b(Y_{x}^{(1)},Y_{x}^{(2)})+1}{2}
\end{multline}
\noindent where the term $\tau_b(\cdot,\cdot)$ is computed based on Eq. (\ref{eq:kendall-tau}).
Here, $\mathcal{D}$ denotes the input dataset consisting of tuples $(x,Y_{x}^{(1)},Y_{x}^{(2)})$.
The mapping operation is equal to assigning a zero-valued weight to the discordant pairs since they do not contribute to the accuracy computation.
\end{definition}

\paragraph{Preference Strength Correlation.}
In human preference modeling, beyond relative ranking, preference strength correlation plays a critical role by quantifying the strength of preferences through continuous numerical values.
However, the current reward-free alignment paradigm often disregards this scalar information, focusing solely on ordinal comparisons.
This omission leads to a loss of preference modeling, which may result in LLMs that fail to accurately capture the subtle variance in human preferences.
Consequently, such models can exhibit suboptimal calibration in generation likelihoods or reward predictions \cite{zhou2024prior}.

\begin{definition}[\textit{Preference Strength Correlation}]
We consider that if an LLM is perfectly aligned with a gold-standard hypothesis space, its indicator function values should exhibit a strong linear correlation with those derived from the gold standard.
To quantify the correlation at the dataset level, we propose an expectation-based Pearson correlation metric, as follows:
\begin{multline}
    \mathrm{PSC}(\mathcal{D}) = 
    \\ \mathbb{E}_{(x,Y_x^{(1)},Y_x^{(2)})\sim\mathcal{D}} \left[ \frac{\mathbb{E}[I_1I_2] - \mathbb{E}[I_1]\mathbb{E}[I_2]}{\sigma_{I_1}\sigma_{I_2}} \right]
\end{multline}
\noindent where the function $\mathbf{I}_1(\cdot,\cdot)$ and $\mathbf{I}_2(\cdot,\cdot)$ represent the indicator functions (generation likelihoods or reward scores) of hypothesis spaces $Y_x^{(1)}$ and $Y_x^{(2)}$ respectively.
Consequently, a well-aligned model would yield a Pearson correlation coefficient approaching 1, reflecting high agreement in preference strength correlation.
\end{definition}

\section{Experiments}
\subsection{Setups}
We evaluated three widely adopted preference optimization algorithms, including DPO \cite{rafailov2024direct}, SimPO \cite{Meng2024SimPOSP}, and ORPO \cite{Hong2024ORPOMP}, using our proposed framework.
For preference annotation and evaluation, we employed ArmoRM-LLaMA-3-8B-v0.1 \cite{Wang2024InterpretablePV} as our primary gold-standard proxy model, ensuring consistency between training and evaluation preference distributions.
To investigate the influence of optimization methods across different preference distributions, we additionally utilized GRM-LLaMA3-8B-rewardmodel-ft \cite{Yang2024RegularizingHS} as a comparative proxy model with a distinct preference distribution.

\begin{table*}[t]
    \centering
    \vspace{-1mm}
    \scalebox{0.76}{
\begin{tabular}{lrrrrrrrrrrrr}
\toprule[1.1pt]
\multicolumn{1}{c}{\multirow{3}{*}{\textbf{\begin{tabular}[c]{@{}c@{}}Model/Method\end{tabular}}}} & \multicolumn{6}{c}{\textbf{w/o Length Normalization}} & \multicolumn{6}{c}{\textbf{w/ Length Normalization}} \\ \cmidrule(r){2-7} \cmidrule(r){8-13}
\multicolumn{1}{c}{} & \multicolumn{2}{c}{\textbf{UniHypo}} & \multicolumn{2}{c}{\textbf{HelpSteer2}} & \multicolumn{2}{c}{\textbf{UltraFeedback}} & \multicolumn{2}{c}{\textbf{UniHypo}} & \multicolumn{2}{c}{\textbf{HelpSteer2}} & \multicolumn{2}{c}{\textbf{UltraFeedback}} \\ \cmidrule(r){2-3} \cmidrule(r){4-5} \cmidrule(r){6-7} \cmidrule(r){8-9} \cmidrule(r){10-11} \cmidrule(r){12-13}    
\multicolumn{1}{c}{} & \multicolumn{1}{r}{\textbf{RA}} & \multicolumn{1}{r}{\textbf{PSC}} & \multicolumn{1}{r}{\textbf{RA}} & \multicolumn{1}{r}{\textbf{PSC}} & \multicolumn{1}{r}{\textbf{RA}} & \multicolumn{1}{r}{\textbf{PSC}} & \multicolumn{1}{r}{\textbf{RA}} & \multicolumn{1}{r}{\textbf{PSC}} & \multicolumn{1}{r}{\textbf{RA}} & \multicolumn{1}{r}{\textbf{PSC}} & \multicolumn{1}{r}{\textbf{RA}} & \multicolumn{1}{r}{\textbf{PSC}} \\ 
\cmidrule(r){1-13}
\multicolumn{13}{l}{\textit{\textbf{Alignment with ArmoRM-Llama3-8B-v0.1 (Same Preference Distribution)}}} \\
\cmidrule(r){1-13}
LLaMA-3.2-3B-Instruct & \underline{54.64} & \underline{0.152} & \textbf{46.79} & \textbf{-0.063} & 53.09 & 0.079 & \underline{48.22} & \underline{-0.048} & 50.69 & 0.013 & \textbf{49.44} & \underline{-0.017} \\
\qquad\qquad\qquad +DPO & \textbf{54.72} & \textbf{0.154} & \textbf{46.79} & \textbf{-0.063} & \underline{53.17} & \textbf{0.081} & \textbf{48.29} & \textbf{-0.046} & \textbf{51.38} & \textbf{0.027} & \underline{49.38} & \textbf{-0.016} \\
\qquad\qquad\qquad +ORPO & 54.55 & 0.151 & \underline{46.68} & \underline{-0.065} & 53.12 & \underline{0.080} & 48.19 & -0.049 & 50.69 & 0.013 & 49.28 & -0.018 \\ 
\qquad\qquad\qquad +SimPO & 54.61 & \underline{0.152} & \underline{46.68} & -0.066 & \textbf{53.22} & \underline{0.080} & 48.21 & \underline{-0.048} & \underline{51.16} & \underline{0.023} & 49.31 & \underline{-0.017} \\ \cmidrule(r){1-13}
LLaMA-3-8B-Instruct & 54.15 & 0.124 & 47.36& -0.051 & 53.13 & 0.079 & \underline{49.81} & \underline{0.031} & 50.81 & 0.016 & 49.86 & -0.011 \\
\qquad\qquad\qquad +DPO & \underline{54.16} & \underline{0.138} & \underline{49.31} & \underline{-0.013} & \underline{64.62} & \underline{0.368} & 47.66 & -0.033 & \underline{55.76} & \underline{0.112} & 59.87 & \underline{0.251} \\
\qquad\qquad\qquad +ORPO & 52.67 & 0.084 & 48.39 & -0.031 & 63.28 & 0.341 & 48.37 & -0.029 & 53.33 & 0.065 & \underline{64.45} & 0.065\\
\qquad\qquad\qquad +SimPO & \textbf{63.59} & \textbf{0.502} & \textbf{53.32} & \textbf{0.065} & \textbf{66.70} & \textbf{0.419} & \textbf{73.30} & \textbf{0.598} & \textbf{66.51} & \textbf{0.319} & \textbf{71.51} & \textbf{0.545} \\
\cmidrule(r){1-13}
\multicolumn{13}{l}{\textit{\textbf{Alignment with GRM-Llama3-8B-rewardmodel-ft (Different Preference Distribution)}}} \\
\cmidrule(r){1-13}

LLaMA-3.2-3B-Instruct & \textbf{51.65} & \textbf{0.059} & \textbf{53.14} & \textbf{0.063} & 50.78 & \underline{0.021} & \textbf{52.08} & \textbf{0.066} & \underline{48.42} & \underline{-0.031} & 51.16 & 0.030 \\
\qquad\qquad\qquad +DPO & \underline{51.64} & \textbf{0.059} & \underline{52.91} & \underline{0.058} & \underline{50.77} & \underline{0.021} & \textbf{52.08} & \textbf{0.066} & \textbf{48.65} & \textbf{-0.027} & \textbf{51.29} & 0.031 \\
\qquad\qquad\qquad +ORPO & \textbf{51.65}& \textbf{0.059} & \textbf{53.14} & \textbf{0.063} & 50.74 & \underline{0.021} & \textbf{52.08} & \textbf{0.066} & \textbf{48.65} & \textbf{-0.027} & \underline{51.21} & \underline{0.032} \\ 
\qquad\qquad\qquad +SimPO &\textbf{51.65} & \textbf{0.059} & \textbf{53.14} & \textbf{0.063} & \textbf{50.84} & \textbf{0.022} & \textbf{52.08} & \textbf{0.066} & 48.19 & -0.036 & \textbf{51.29} & \textbf{0.033} \\ \cmidrule(r){1-13}
LLaMA-3-8B-Instruct & \textbf{51.68} & \textbf{0.056} & 52.58 & 0.051 & 49.92 & 0.008 & \textbf{52.67} & \textbf{0.076} & 48.53 & -0.029 & 50.13 & \underline{0.013} \\
\qquad\qquad\qquad +DPO & 51.29 & 0.049 & \underline{54.48} & \underline{0.089} & \textbf{50.54} & \textbf{0.014} & 51.78 & 0.054 & 47.52 & -0.049 & \underline{50.31} & 0.007 \\
\qquad\qquad\qquad +ORPO & \underline{51.56} & \underline{0.053} & 52.47 & 0.049 & 50.10 & 0.009 & \underline{52.42} & \underline{0.072} & \textbf{49.66} & \textbf{-0.007} & 50.01 & 0.007 \\
\qquad\qquad\qquad +SimPO & 49.50 & -0.031 & \textbf{55.03} & \textbf{0.100} & \underline{50.46} & \underline{0.013} & 50.73 & -0.008 & \underline{48.98} & \underline{-0.020} & \textbf{50.99} & \textbf{0.023} \\
\bottomrule[1.1pt]
\end{tabular}}
    \caption{
        Experimental results on different preference optimization methods.
        RA and PSC denote ranking accuracy and preference strength correlation, respectively.
        The best results for each group are in \textbf{bold}.
        The second-best results for each group are with \underline{underline}.
    }
    \vspace{-0.4cm}
    \label{tab:main-result}
\end{table*}

\paragraph{Datasets.}
We employed the following datasets for training and evaluation:

(1) \textbf{UltraFeedback} \cite{Cui2023UltraFeedbackBL}: 
A large-scale preference dataset comprising 64k prompts and 256k responses.
We performed preference optimization on the training split and utilized its validation set for in-distribution evaluation.

(2) \textbf{HelpSteer2-Preference} \cite{Wang2024HelpSteer2PreferenceCR}:
A high-quality dataset annotated with preference directions, strength scores, and textual justifications.
Similarly, we conducted the evaluation on its validation split.

(3) \textbf{UniHypoBench}:
To address the limitation of the existing evaluation sets (which typically provide less than 4 responses per prompt), we constructed the Unified Hypothesis Benchmark (UniHypoBench).
Curated from RewardBench \cite{lambert2024rewardbench}, it extends the instruction set with 2,985 prompts, each containing more than 8 responses sampled from diverse LLMs, enabling more comprehensive analysis.

We construct the benchmark by first collecting hypothesis response samples from multiple state-of-the-art commercial and open-source LLMs, as detailed in Appendix \ref{app:construct-unihypo}. 
To ensure optimal response diversity while preserving output quality, we configure the generation process with a temperature of 0.75 and top-p sampling at 0.95, as well a 768-token truncation limit for all responses.

\paragraph{Models.}
We evaluated our approach using three models, including LLaMA-3.2-3B-Instruct and LLaMA-3-8B-Instruct.
For the LLaMA-3.2-3B-Instruct model base, we conducted preference optimizations.
Concurrently, the LLaMA-3-8B-Instruct models were evaluated using the pre-optimized weights released by \citet{Meng2024SimPOSP}.

\paragraph{Training Settings.} 
We conducted preference optimization using an effective batch size of 128 and a maximum sequence length of 1024.
The learning rate follows a cosine decay schedule with 10\% warmup steps over one training epoch.
For method-specific hyperparameters, we performed a grid search to determine the optimal configuration.



\subsection{Main Results}
\label{sec:main-results}
We conduct an evaluation of diverse preference learning methods using our proposed framework.
To ensure the preference consistency, we maintain identical rating proxy models between training and evaluation phases, thereby guaranteeing that all compared methods learn from and are assessed against the same preference distribution.
Additionally, we apply the original instruction models as baselines, which enables a quantitative assessment of the performance gains achieved through explicit preference learning.
We present the main results in Table \ref{tab:main-result}.
The results demonstrate:


\paragraph{\textbf{Preference Optimization Effectively Captures the Preference Information.}}
Our results show that preference optimization methods generally outperform baselines in both ranking accuracy and preference strength correlation, confirming their effectiveness in capturing preference distributions.
The LLaMA-3-8B-Instruct model benefits most significantly, with SimPO achieving over 10\% improvement across all datasets.
However, even SimPO's best performance (66.70\% on UltraFeedback) remains suboptimal, aligning with \citeauthor{Chen2024PreferenceLA}'s observation that current methods still have substantial room for improvement.




\vspace{-0.1cm}
\paragraph{\textbf{Preference Optimization Learns Model-Specific Preference Patterns.}}
Our evaluation reveals that while preference optimization improves alignment with the training proxy model, these gains often fail to generalize to other proxy models with different distributions.
In some cases, we even observe performance degradation when evaluating against alternative proxies.
These findings demonstrate that current methods primarily learn model-specific judging patterns rather than general preferences.
This specificity poses a fundamental challenge for the LLM-based evaluation, as different evaluators may employ conflicting preference criteria, complicating the assessment of alignment quality.

\begin{figure*}
    \centering
    \subfigure[ID Preference Overlaps]{
        \includegraphics[scale=0.45]{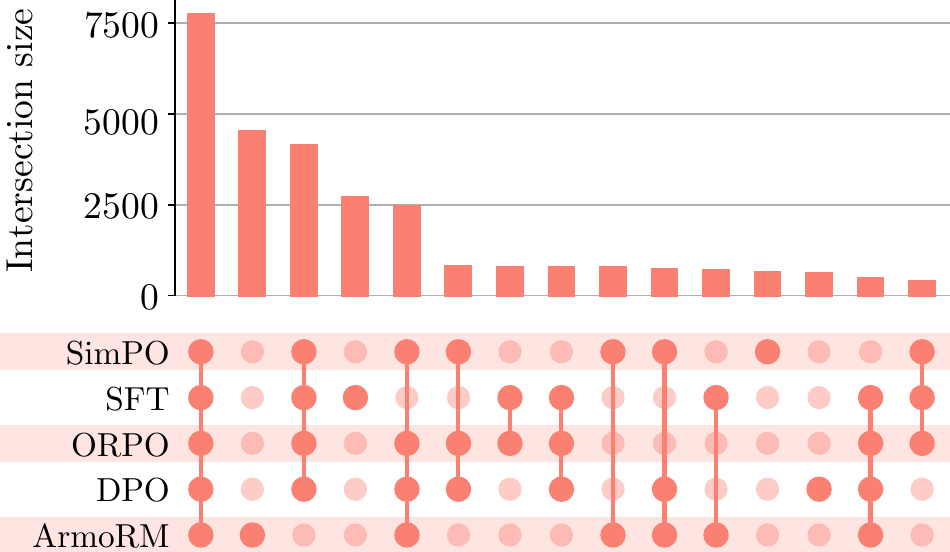}
    }
    \subfigure[OOD Preference Overlaps]{
        \includegraphics[scale=0.45]{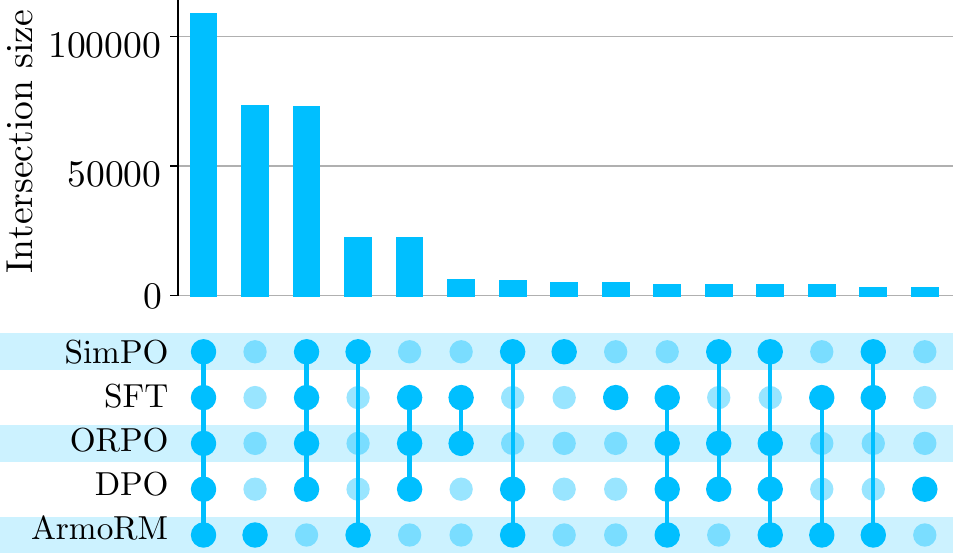}
    }
    \vspace{-0.3cm}
    \caption{
    Upset plots of preference intersections on the OOD test set (UniHypo).
    The upper bar chart displays the amount of preference overlaps between different methods, while the lower connection matrix identifies the constituent subsets of each intersection. Full results can be found in Figure \ref{fig:upset-plots-w-ln}.
    }
    \label{fig:upset-plot}
    \vspace{-0.3cm}
\end{figure*}

\vspace{-0.1cm}
\paragraph{\textbf{Length Normalization is Potential in Preference Modeling.}}
Our analysis indicates that length normalization generally impairs ranking accuracy. However, the normalized SimPO version of LLaMA-3-8B-Instruct achieves a 73.3\% accuracy, surpassing the performance of unnormalized models.
This demonstrates that length-aware objectives can learn better preference representations, suggesting their value for future methods.

\vspace{-0.1cm}
\paragraph{\textbf{Capturing Subtle Preference Correlation is Challenging.}}

Current alignment methods exhibit strong ranking accuracy but exhibit a weak correlation with preference strength, typically below 0.3. This result highlights the challenges in quantifying preference strength. However, SimPO stands out as an exception, achieving a correlation of 0.419 on UltraFeedback (up from 0.079), demonstrating that improved strength modeling is achievable.

\subsection{Analysis}
\paragraph{\textbf{Visualization of Preference Intersections.}}

We employ upset plots  \cite{2014_infovis_upset} to analyze preference intersections across hypothesis spaces (Figure \ref{fig:upset-plot}), presenting both in-distribution (a) and out-of-distribution (b) results.
For clarity, we focus on the plot's forepart, where solid-connected points mark shared preference tendencies across response pairs.
Our key observations of the in-distribution test set include:
(1) The largest intersection demonstrates fundamental preference knowledge shared by all optimization methods and the proxy model.
(2) The substantial second intersection indicates significant unlearned preferences.
(3) The fourth intersection shows that methods successfully capture novel preferences from the proxy model with notable behavioral deviation.

In parallel, we conduct the identical visualization on the out-of-distribution test set.
Apart from the observations in Figure \ref{fig:upset-plot} (a), we find that:
(1) Unlearned preferences increase proportionally, revealing domain-shift effects.
(2) While most methods degrade, SimPO maintains the largest intersection, demonstrating superior generalization.
(3) The overall performance decline underscores the need for more robust preference learning paradigms.
These visual analyses provide intuitive mechanistic insights that corroborate our quantitative findings in Section \ref{sec:main-results}.

\paragraph{\textbf{Alignment in LLM's Internal Preference Distribution.}}
To gain deeper insights into the alignment effects, we analyze the internal preference distribution using UniHypoBench.
We sample responses from aligned LLaMA-3-8B-Instruct-based models (with the SFT base model as baseline) at a temperature of 0.75 to ensure sufficient diversity.
The generation likelihoods are then extracted to compute both ranking accuracy and preference strength correlation, as listed in Table \ref{tab:self-dist}.
Surprisingly, the results reveal that performance shows no significant improvement even when evaluated on the model's own preference distribution.
More notably, we observe performance degradation in some cases, particularly for the SimPO-aligned model.
We assume that this phenomenon probably stems from the scarcity of the diversity of these sampled hypotheses.
This finding also notes that distinguishing the subtle difference between similar hypotheses is a challenge for further development of 
the preference learning method.

\begin{table}[t]
    \centering
    \scalebox{0.91}{
\begin{tabular}{lrrrr}
\toprule[1.1pt]
\multicolumn{1}{c}{Metric/Method} & \multicolumn{1}{c}{SFT} & \multicolumn{1}{c}{DPO} & \multicolumn{1}{c}{ORPO} & \multicolumn{1}{c}{SimPO} \\
\cmidrule(r){1-5}
RA w/o LN & 55.93 & 55.99 & 55.01 & \textbf{61.50} \\
PSC w/o LN & 0.164 & 0.153 & 0.123 & \textbf{0.301} \\
\cmidrule(r){1-5}
RA w/ LN & 54.12 & 51.49 & 54.74 & \textbf{59.70} \\
PSC w/ LN & 0.100 & 0.044 & 0.108 & \textbf{0.252} \\
\bottomrule[1.1pt]
\end{tabular}}
    \caption{
        Experimental results of LLaMA-3-8B-Instruct-based models' internal preference distribution.
        The best results for each group are in \textbf{bold}.
        LN denotes length normalization.
    }
    \vspace{-0.4cm}
    \label{tab:self-dist}
\end{table}

\begin{figure*}
    \centering
    \subfigure[Internal Preference Distribution]{
        \scalebox{0.7}{\includegraphics[scale=0.43]{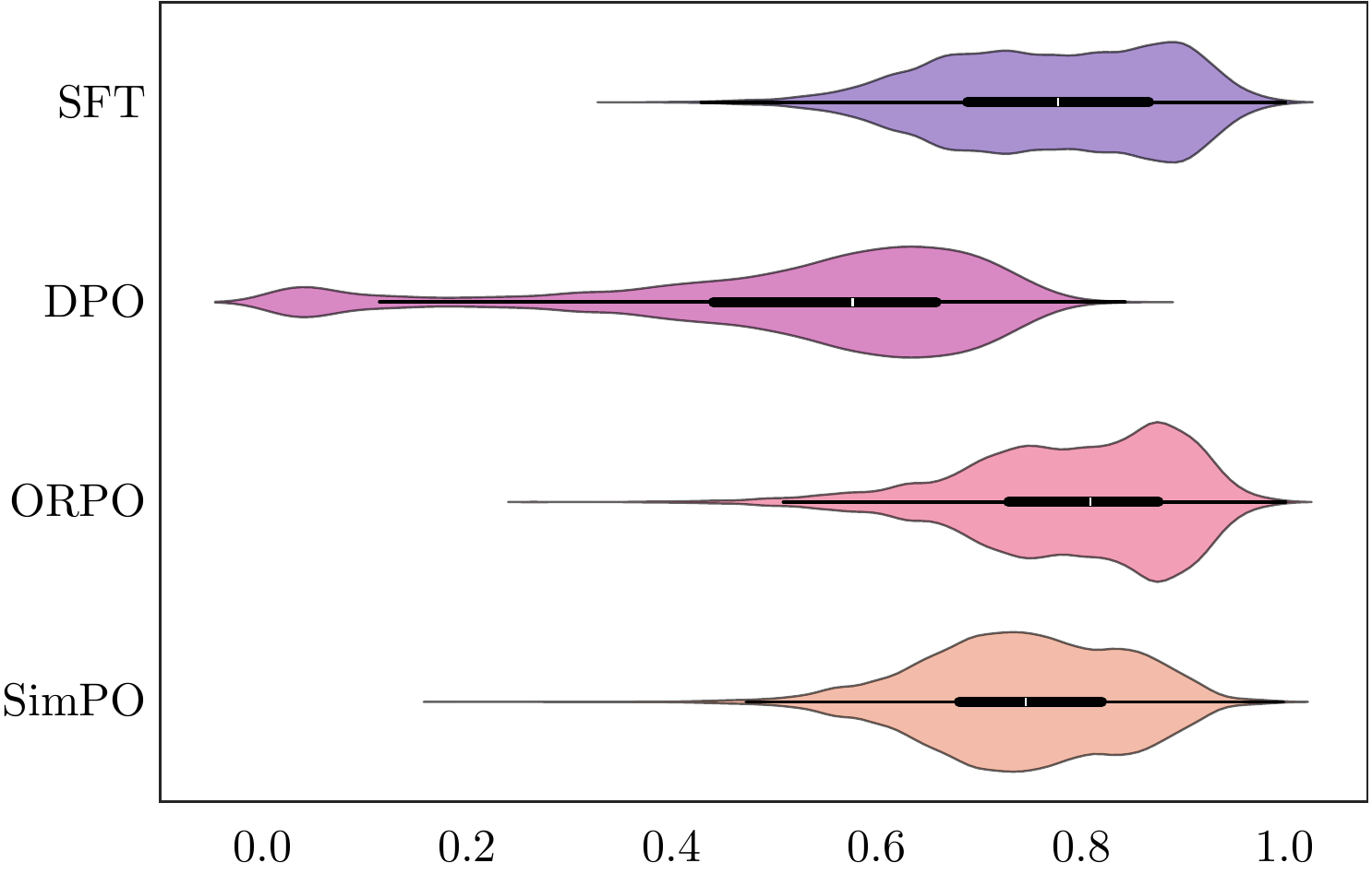}}
    }
    \subfigure[External Preference Distribution]{
        \scalebox{0.7}{\includegraphics[scale=0.43]{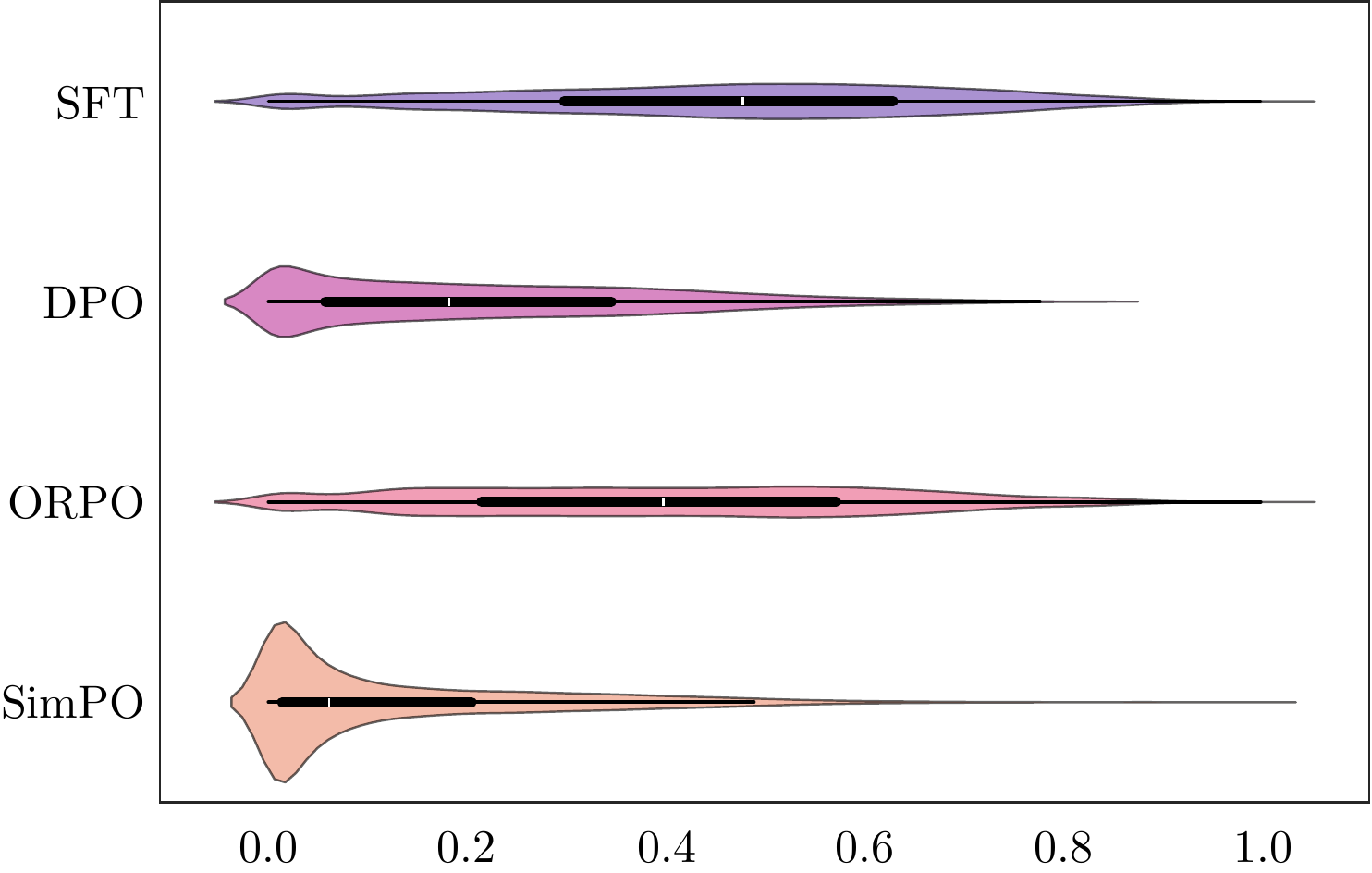}}
    }
    \vspace{-0.3cm}
    \caption{
    Violin plots comparing generation likelihood distributions across different alignment methods (including the SFT base model as baseline).
    The plot width represents probability density, with the central white line marking the median value for each distribution.
    To enhance the readability of the chart, we employ length normalization.
    }
    \vspace{-0.3cm}
    \label{fig:violin-plot}
\end{figure*}

To gain deeper insight into the mechanisms of preference learning, we conduct a density-based analysis of generation likelihood distributions. 
Figure \ref{fig:violin-plot} presents violin plots of these distributions across different alignment approaches, with the density curves estimated using kernel density methods.
As shown in Figure \ref{fig:violin-plot} (a), which visualizes the internal preference distributions, we observe remarkably consistent patterns across all optimization methods. 
The finding indicates that these optimization methods probably do not vary much in their internal preference distributions.


Our analysis of external preference distributions in Figure \ref{fig:violin-plot} (b) reveals distinct patterns that contrast with the internal consistency observed previously.
The distributions exhibit a clear dichotomy: while SFT and ORPO maintain near-uniform distributions, DPO and SimPO obtain spindle-shaped distributions, reflecting their enhanced capability to suppress negative samples through preference optimization. 
This successful suppression of undesirable outputs represents a significant advancement in alignment techniques.

However, closer examination reveals a critical limitation - none of the methods achieve the theoretically optimal bimodal distribution that would fully separate preferred and rejected responses. 
This persistent unimodality suggests that while current approaches can effectively downweight negative samples, they struggle to develop truly discriminative representations that clearly partition the hypothesis space. 
The gap between empirical results and theoretical expectations points to fundamental constraints in existing optimization frameworks, which appear to learn primarily through global likelihood adjustment rather than developing more sophisticated, robust representations of preference structure.

\section{Discussion}
\label{sec:discussion}
\paragraph{\textbf{Using Human Annotation as a Proxy Model.}}

While the previous sections demonstrated how HEAL measures alignment with a single \textit{gold-standard} proxy, a key advantage of our framework is its ability to perform a unified evaluation across a diverse \textit{hypothesis space}. 
The ranking-based structure of this space allows HEAL to seamlessly incorporate gold standards from various sources, including generative language models, reward models, and human annotators.
We further explore the distinction between reward models and human judgment by incorporating the original human annotations from HelpSteer2, with the results shown in Table \ref{tab:human-comparsion} details the findings from this experiment.

\begin{table*}[t]
    \centering
    \scalebox{0.92}{
\begin{tabular}{lrrrrrrrrrr}
\toprule[1.1pt]
\multirow{2}{*}{Model/Method} & \multicolumn{2}{c}{Helpfulness} & \multicolumn{2}{c}{Correctness} & \multicolumn{2}{c}{Coherence} & \multicolumn{2}{c}{Complexity} & \multicolumn{2}{c}{Verbosity} \\
 & RA. & PSC. & RA. & PSC. & RA. & PSC. & RA. & PSC. & RA. & PSC. \\
\cmidrule(r){1-11}
LLaMA-3-8B-Instruct & 46.61 & -0.068 & 45.25 & -0.095 & 50.44 & 0.009 & \textbf{32.14} & \textbf{-0.357} & \textbf{26.87} & \textbf{-0.463} \\
\qquad\qquad\qquad +DPO & 47.25 & -0.055 & 45.93 & -0.081 & 52.86 & 0.057 & 31.59 & -0.368 & 23.47 & -0.531 \\
\qquad\qquad\qquad +SimPO & 44.79 & -0.104 & 44.13 & -0.117 & \textbf{54.82} & \textbf{0.096} & 29.55 & -0.409 & 23.13 & -0.537 \\
\qquad\qquad\qquad +ORPO & \textbf{48.18} & \textbf{-0.037} & \textbf{46.93} & \textbf{-0.061} & 51.32 & 0.026 & \textbf{32.14} & \textbf{-0.357} & 24.15 & -0.517 \\
\bottomrule[1.1pt]
\end{tabular}}
    \caption{
        Experimental results on different preference dimensions.
    }
    \vspace{-0.4cm}
    \label{tab:multidimension}
\end{table*}

\begin{table}[t]
    \centering
    \scalebox{0.88}{


\resizebox{0.54\textwidth}{!}{\begin{tabular}{lrrrr}
\toprule[1.1pt]
\multirow{2}{*}{Model/Method} & \multicolumn{2}{c}{Armo-RM} & \multicolumn{2}{c}{Human Annotation} \\
 & RS. & PSC. & RS. & PSC. \\
 \cmidrule(r){1-5}
LLaMA-3-8B-Instruct & \textbf{66.79} & \textbf{0.336} & 46.61 & -0.068 \\
\qquad\qquad\qquad +DPO & 49.70 & -0.006 & 47.25 & -0.055 \\
\qquad\qquad\qquad +SimPO & 47.10 & -0.058 & 44.79 & -0.104 \\
\qquad\qquad\qquad +ORPO & 60.03 & 0.201 & \textbf{48.18} & \textbf{-0.037} \\
\bottomrule[1.1pt]
\end{tabular}}
}
    \caption{
        Experimental results on human annotation.
    }
    \vspace{-0.4cm}
    \label{tab:human-comparsion}
\end{table}

We observe that preference optimization demonstrates stronger alignment with our proxy reward model than with human annotations. 
Then, despite this quantitative difference, both evaluation methods reveal resemblance in their assessments, indicating correlation between reward model and human evaluation paradigms. 
This empirical evidence suggests that reward models can serve as a computationally efficient proxy for alignment assessment, particularly when resource constraints make large-scale human evaluation impractical.

\paragraph{\textbf{A Multidimensional Study on Preferences.}}
The complex, multidimensional nature of human preferences makes alignment evaluation challenging under the standard \textit{LLM-as-a-Judge} paradigm.
In contrast, HEAL is specifically designed for this complexity, offering the flexibility to assess preferences along any given sub-dimension. 
We expand our analysis to incorporate comprehensive multidimensional assessment.
We demonstrate this capability through a multidimensional analysis of the HelpSteer2 dataset, which provides human annotations across five key dimensions: \texttt{helpfulness}, \texttt{correctness}, \texttt{coherence}, \texttt{complexity}, and \texttt{verbosity} in Table \ref{tab:multidimension}, our findings reveal that preference learning algorithms are notably adept at capturing most of these sub-dimensions. 

\paragraph{\textbf{Distinction Existing Evaluation Methods.}}
HEAL distinguishes itself from traditional pairwise ranking evaluations (\textit{e.g.}, RewardBench, \textit{LLM-as-a-Judge}) by not being an incremental improvement on existing methods. 
Instead, HEAL utilizes ranking as just one component within a broader analytical structure. 
To the best of our knowledge, HEAL is the first framework to adopt a \textit{hypothesis-driven} perspective for alignment evaluation. 
It advances prior approaches by introducing a unified framework that seamlessly integrates generative models, reward models, and human feedback. 
This distinctive architecture enables profound flexibility in alignment measurement—a core contribution of our work. 
We further detail these key differences in Appendix \ref{sec:relation-exist-work}.


\section{Conclusion}

This paper presented HEAL, a novel, \textit{hypothesis-based} framework designed to move beyond traditional evaluation of preference learning. 
By employing two complementary metrics, HEAL provides a more holistic assessment of model alignment.
Our evaluation using HEAL yields three key observations in current models, revealing suboptimal ranking performance, an inability to capture preference nuances, and unique, model-specific learning behaviors.
Ultimately, HEAL offers the community a more powerful and nuanced tool for diagnosing and improving preference learning, paving the way for the development of more reliably aligned models.

\pagebreak

\section*{Acknowledgments}

This work was supported in part by the National Science Foundation of China (Nos. 62276056 and U24A20334), the Yunnan Fundamental Research Projects (No.202401BC070021), the Yunnan Science and Technology Major Project (No. 202502AD080014), and the Program of Introducing Talents of Discipline to Universities, Plan 111 (No.B16009).

\section*{Limitations}

While our proposed HEAL framework provides a novel hypothesis-based approach for preference-aware analysis, several limitations require discussion. 
First, our experimental validation, though demonstrating practical utility for resource-constrained scenarios, was conducted on a limited set of models, with LLaMA-3-8B-Instruct serving as the primary exemplar due to its consistently strong performance. 
Second, while ranking accuracy and preference strength correlation prove effective as evaluation metrics, future work may identify more sophisticated measures that better capture the nuances of preference learning. 
Finally, our current analysis does not examine the training dynamics of these metrics during optimization, leaving open questions about their evolution and relationship to model convergence. 
These limitations point to valuable directions for future research, particularly in developing more comprehensive analysis approaches and investigating the inherent mechanism of preference alignment.

\section*{Ethics Statement}

This work does not need ethical considerations.
Although in this work we construct data as described in Appendix \ref{app:construct-unihypo}, this input is all from open-source data, and the output is also obtained based on open-source or commercial models.

\bibliography{acl_latex}

\begin{thebibliography}{33}
\providecommand{\natexlab}[1]{#1}

\bibitem[{Bai et~al.(2024)Bai, Liu, Bu, He, Liu, Zhou, Lin, Su, Ge, Zheng, and Ouyang}]{Bai2024MTBench101AF}
Ge~Bai, Jie Liu, Xingyuan Bu, Yancheng He, Jiaheng Liu, Zhanhui Zhou, Zhuoran Lin, Wenbo Su, Tiezheng Ge, Bo~Zheng, and Wanli Ouyang. 2024.
\newblock \href {https://arxiv.org/abs/2402.14762} {Mt-bench-101: A fine-grained benchmark for evaluating large language models in multi-turn dialogues}.
\newblock \emph{ArXiv preprint}, abs/2402.14762.

\bibitem[{Bai et~al.(2022)Bai, Jones, Ndousse, Askell, Chen, DasSarma, Drain, Fort, Ganguli, Henighan et~al.}]{bai2022training}
Yuntao Bai, Andy Jones, Kamal Ndousse, Amanda Askell, Anna Chen, Nova DasSarma, Dawn Drain, Stanislav Fort, Deep Ganguli, Tom Henighan, and 1 others. 2022.
\newblock Training a helpful and harmless assistant with reinforcement learning from human feedback.
\newblock \emph{arXiv preprint arXiv:2204.05862}.

\bibitem[{Chen et~al.(2024)Chen, Malladi, Zhang, Chen, Zhang, Ranganath, and Cho}]{Chen2024PreferenceLA}
Angelica Chen, Sadhika Malladi, Lily~H. Zhang, Xinyi Chen, Qiuyi Zhang, Rajesh Ranganath, and Kyunghyun Cho. 2024.
\newblock \href {https://api.semanticscholar.org/CorpusID:270123277} {Preference learning algorithms do not learn preference rankings}.
\newblock \emph{ArXiv}, abs/2405.19534.

\bibitem[{Cui et~al.(2023)Cui, Yuan, Ding, Yao, Zhu, Ni, Xie, Liu, and Sun}]{Cui2023UltraFeedbackBL}
Ganqu Cui, Lifan Yuan, Ning Ding, Guanming Yao, Wei Zhu, Yuan Ni, Guotong Xie, Zhiyuan Liu, and Maosong Sun. 2023.
\newblock \href {https://arxiv.org/abs/2310.01377} {Ultrafeedback: Boosting language models with high-quality feedback}.
\newblock \emph{ArXiv preprint}, abs/2310.01377.

\bibitem[{Dubey et~al.(2024)Dubey, Jauhri, Pandey, Kadian, Al-Dahle, Letman, Mathur, Schelten, Yang, Fan, Goyal, Hartshorn, Yang, Mitra, Sravankumar, Korenev, Hinsvark, Rao, Zhang, Rodriguez, Gregerson, Spataru, Rozi{\`e}re, Biron, Tang, Chern, Caucheteux, Nayak, Bi, Marra, McConnell, Keller, Touret, Wu, Wong, tian Cant{\'o}n~Ferrer, Nikolaidis, Allonsius, Song, Pintz, Livshits, Esiobu, Choudhary, Mahajan, Garcia-Olano, Perino, Hupkes, Lakomkin, AlBadawy, Lobanova, Dinan, Smith, Radenovic, Zhang, Synnaeve, Lee, Anderson, Nail, Mialon, Pang, Cucurell, Nguyen, Korevaar, Xu, Touvron, Zarov, Ibarra, Kloumann, Misra, Evtimov, Copet, Lee, Geffert, Vranes, Park, Mahadeokar, Shah, van~der Linde, Billock, Hong, Lee, Fu, Chi, Huang, Liu, Wang, Yu, Bitton, Spisak, Park, Rocca, Johnstun, Saxe, Jia, Alwala, Upasani, Plawiak, Li, neth Heafield, Stone, El-Arini, Iyer, Malik, ley Chiu, Bhalla, Rantala-Yeary, van~der Maaten, Chen, Tan, Jenkins, Martin, Madaan, Malo, Blecher, Landzaat, de~Oliveira, Muzzi, Pasupuleti, Singh,
  Paluri, Kardas, Oldham, Rita, Pavlova, Kambadur, Lewis, Si, Singh, Hassan, Goyal, Torabi, lay Bashlykov, Bogoychev, Chatterji, Duchenne, cCelebi, Alrassy, Zhang, Li, Vasi{\'c}, Weng, Bhargava, Dubal, Krishnan, Koura, Xu, He, Dong, Srinivasan, Ganapathy, Calderer, Cabral, Stojnic, Raileanu, Girdhar, Patel, main Sauvestre, nie Polidoro, Sumbaly, Taylor, Silva, Hou, Wang, Hosseini, hana Chennabasappa, Singh, Bell, Kim, Edunov, Nie, Narang, Raparthy, Shen, Wan, Bhosale, Zhang, Vandenhende, Batra, Whitman, Sootla, Collot, Gururangan, Borodinsky, Herman, Fowler, Sheasha, Georgiou, Scialom, Speckbacher, Mihaylov, Xiao, Karn, Goswami, Gupta, Ramanathan, Kerkez, Gonguet, ginie Do, Vogeti, Petrovic, Chu, Xiong, Fu, ney Meers, Martinet, Wang, Tan, Xie, Jia, Wang, Goldschlag, Gaur, Babaei, Wen, Song, Zhang, Li, Mao, Coudert, Yan, Chen, Papakipos, Singh, Grattafiori, Jain, Kelsey, Shajnfeld, Gangidi, Victoria, Goldstand, Menon, Sharma, Boesenberg, Vaughan, Baevski, Feinstein, Kallet, Sangani, Yunus, Lupu, Alvarado,
  Caples, Gu, Ho, Poulton, Ryan, Ramchandani, Franco, Saraf, Chowdhury, Gabriel, Bharambe, Eisenman, Yazdan, James, Maurer, Leonhardi, Huang, Loyd, de~Paola, Paranjape, Liu, Wu, Ni, Hancock, Wasti, Spence, Stojkovic, Gamido, Montalvo, Parker, Burton, Mejia, Wang, Kim, Zhou, Hu, Chu, Cai, Tindal, Feichtenhofer, mon Civin, Beaty, Kreymer, Li, Wyatt, Adkins, Xu, Testuggine, David, Parikh, Liskovich, Foss, Wang, Le, Holland, Dowling, Jamil, Montgomery, Presani, Hahn, Wood, Brinkman, Arcaute, Dunbar, Smothers, Sun, Kreuk, Tian, Ozgenel, Caggioni, Guzm’an, Kanayet, Seide, Florez, Schwarz, Badeer, Swee, Halpern, Thattai, Herman, Sizov, Zhang, Lakshminarayanan, Shojanazeri, Zou, Wang, Zha, Habeeb, Rudolph, Suk, Aspegren, Goldman, Molybog, Tufanov, Veliche, Gat, Weissman, Geboski, Kohli, Asher, Gaya, Marcus, Tang, Chan, Zhen, Reizenstein, Teboul, Zhong, Jin, Yang, Cummings, Carvill, Shepard, McPhie, Torres, Ginsburg, Wang, Wu, KamHou, Saxena, Prasad, Khandelwal, Zand, Matosich, Veeraraghavan, Michelena, Li, Huang,
  Chawla, Lakhotia, Huang, Chen, Garg, Lavender, Silva, Bell, Zhang, Guo, Yu, Moshkovich, Wehrstedt, Khabsa, Avalani, Bhatt, Tsimpoukelli, Mankus, Hasson, Lennie, Reso, Groshev, Naumov, Lathi, Keneally, Seltzer, Valko, Restrepo, Patel, Vyatskov, Samvelyan, Clark, Macey, Wang, Hermoso, Metanat, Rastegari, Bansal, Santhanam, Parks, White, ata Bawa, Singhal, Egebo, Usunier, Laptev, Dong, Zhang, Cheng, Chernoguz, Hart, Salpekar, Kalinli, Kent, Parekh, Saab, Balaji, dro Rittner, Bontrager, Roux, Doll{\'a}r, Zvyagina, Ratanchandani, Yuvraj, Liang, Alao, Rodriguez, Ayub, Murthy, Nayani, Mitra, Li, Hogan, Battey, Wang, Maheswari, Howes, Rinott, Bondu, Datta, Chugh, Hunt, Dhillon, Sidorov, Pan, Verma, Yamamoto, Ramaswamy, Lindsay, Feng, Lin, Zha, Shankar, Zhang, Wang, Agarwal, Sajuyigbe, Chintala, Max, Chen, Kehoe, Satterfield, Govindaprasad, Gupta, Cho, Virk, Subramanian, Choudhury, Goldman, Remez, Glaser, Best, Kohler, Robinson, Li, Zhang, Matthews, Chou, Shaked, Vontimitta, Ajayi, Montanez, Mohan, Kumar, Mangla,
  Ionescu, Poenaru, Mihailescu, Ivanov, Li, Wang, Jiang, Bouaziz, Constable, Tang, Wang, Wu, Wang, Xia, Wu, Gao, Chen, Hu, Jia, Qi, Li, Zhang, Zhang, Adi, Nam, Wang, Hao, Qian, He, Rait, DeVito, Rosnbrick, Wen, Yang, and Zhao}]{Dubey2024TheL3}
Abhimanyu Dubey, Abhinav Jauhri, Abhinav Pandey, Abhishek Kadian, Ahmad Al-Dahle, Aiesha Letman, Akhil Mathur, Alan Schelten, Amy Yang, Angela Fan, Anirudh Goyal, Anthony~S. Hartshorn, Aobo Yang, Archi Mitra, Archie Sravankumar, Artem Korenev, Arthur Hinsvark, Arun Rao, Aston Zhang, and 510 others. 2024.
\newblock \href {https://api.semanticscholar.org/CorpusID:271571434} {The llama 3 herd of models}.
\newblock \emph{ArXiv}, abs/2407.21783.

\bibitem[{Dubois et~al.(2024)Dubois, Galambosi, Liang, and Hashimoto}]{Dubois2024LengthControlledAA}
Yann Dubois, Bal'azs Galambosi, Percy Liang, and Tatsunori Hashimoto. 2024.
\newblock \href {https://arxiv.org/abs/2404.04475} {Length-controlled alpacaeval: A simple way to debias automatic evaluators}.
\newblock \emph{ArXiv preprint}, abs/2404.04475.

\bibitem[{Ethayarajh et~al.(2024)Ethayarajh, Xu, Muennighoff, Jurafsky, and Kiela}]{Ethayarajh2024KTOMA}
Kawin Ethayarajh, Winnie Xu, Niklas Muennighoff, Dan Jurafsky, and Douwe Kiela. 2024.
\newblock \href {https://api.semanticscholar.org/CorpusID:267406810} {Kto: Model alignment as prospect theoretic optimization}.
\newblock \emph{ArXiv}, abs/2402.01306.

\bibitem[{Gao et~al.(2022)Gao, Schulman, and Hilton}]{Gao2022ScalingLF}
Leo Gao, John Schulman, and Jacob Hilton. 2022.
\newblock \href {https://api.semanticscholar.org/CorpusID:252992904} {Scaling laws for reward model overoptimization}.
\newblock In \emph{International Conference on Machine Learning}.

\bibitem[{Gu et~al.(2024)Gu, Jiang, Shi, Tan, Zhai, Xu, Li, Shen, Ma, Liu, Wang, and Guo}]{Gu2024ASO}
Jiawei Gu, Xuhui Jiang, Zhichao Shi, Hexiang Tan, Xuehao Zhai, Chengjin Xu, Wei Li, Yinghan Shen, Shengjie Ma, Honghao Liu, Yuanzhuo Wang, and Jian Guo. 2024.
\newblock \href {https://api.semanticscholar.org/CorpusID:274234014} {A survey on llm-as-a-judge}.
\newblock \emph{ArXiv}, abs/2411.15594.

\bibitem[{Hong et~al.(2024)Hong, Lee, and Thorne}]{Hong2024ORPOMP}
Jiwoo Hong, Noah Lee, and James Thorne. 2024.
\newblock \href {https://api.semanticscholar.org/CorpusID:268363309} {Orpo: Monolithic preference optimization without reference model}.
\newblock \emph{ArXiv}, abs/2403.07691.

\bibitem[{Jiang et~al.(2023)Jiang, Liu, Liu, Zhao, Zhang, Gao, Zhang, Li, and Xiong}]{jiang2023clip}
Dongsheng Jiang, Yuchen Liu, Songlin Liu, Jin'e Zhao, Hao Zhang, Zhen Gao, Xiaopeng Zhang, Jin Li, and Hongkai Xiong. 2023.
\newblock From clip to dino: Visual encoders shout in multi-modal large language models.
\newblock \emph{arXiv preprint arXiv:2310.08825}.

\bibitem[{Kwon et~al.(2023)Kwon, Li, Zhuang, Sheng, Zheng, Yu, Gonzalez, Zhang, and Stoica}]{Kwon2023EfficientMM}
Woosuk Kwon, Zhuohan Li, Siyuan Zhuang, Ying Sheng, Lianmin Zheng, Cody~Hao Yu, Joseph~E. Gonzalez, Haotong Zhang, and Ion Stoica. 2023.
\newblock \href {https://api.semanticscholar.org/CorpusID:261697361} {Efficient memory management for large language model serving with pagedattention}.
\newblock \emph{Proceedings of the 29th Symposium on Operating Systems Principles}.

\bibitem[{Lambert et~al.(2024)Lambert, Pyatkin, Morrison, Miranda, Lin, Chandu, Dziri, Kumar, Zick, Choi et~al.}]{lambert2024rewardbench}
Nathan Lambert, Valentina Pyatkin, Jacob Morrison, LJ~Miranda, Bill~Yuchen Lin, Khyathi Chandu, Nouha Dziri, Sachin Kumar, Tom Zick, Yejin Choi, and 1 others. 2024.
\newblock \href {https://arxiv.org/abs/2403.13787} {Rewardbench: Evaluating reward models for language modeling}.
\newblock \emph{ArXiv preprint}, abs/2403.13787.

\bibitem[{Lex et~al.(2014)Lex, Gehlenborg, Strobelt, Vuillemot, and Pfister}]{2014_infovis_upset}
Alexander Lex, Nils Gehlenborg, Hendrik Strobelt, Romain Vuillemot, and Hanspeter Pfister. 2014.
\newblock \href {https://doi.org/10.1109/TVCG.2014.2346248} {Upset: Visualization of intersecting sets}.
\newblock \emph{IEEE Transactions on Visualization and Computer Graphics (InfoVis)}, 20(12):1983--1992.

\bibitem[{Li et~al.(2024)Li, Chiang, Frick, Dunlap, Wu, Zhu, Gonzalez, and Stoica}]{Li2024FromCD}
Tianle Li, Wei-Lin Chiang, Evan Frick, Lisa Dunlap, Tianhao Wu, Banghua Zhu, Joseph Gonzalez, and Ion Stoica. 2024.
\newblock \href {https://api.semanticscholar.org/CorpusID:270562889} {From crowdsourced data to high-quality benchmarks: Arena-hard and benchbuilder pipeline}.
\newblock \emph{ArXiv}, abs/2406.11939.

\bibitem[{Meng et~al.(2024)Meng, Xia, and Chen}]{Meng2024SimPOSP}
Yu~Meng, Mengzhou Xia, and Danqi Chen. 2024.
\newblock \href {https://arxiv.org/abs/2405.14734} {Simpo: Simple preference optimization with a reference-free reward}.
\newblock \emph{ArXiv preprint}, abs/2405.14734.

\bibitem[{Ouyang et~al.(2022)Ouyang, Wu, Jiang, Almeida, Wainwright, Mishkin, Zhang, Agarwal, Slama, Ray, Schulman, Hilton, Kelton, Miller, Simens, Askell, Welinder, Christiano, Leike, and Lowe}]{Ouyang2022TrainingLM}
Long Ouyang, Jeff Wu, Xu~Jiang, Diogo Almeida, Carroll~L. Wainwright, Pamela Mishkin, Chong Zhang, Sandhini Agarwal, Katarina Slama, Alex Ray, John Schulman, Jacob Hilton, Fraser Kelton, Luke~E. Miller, Maddie Simens, Amanda Askell, Peter Welinder, Paul~Francis Christiano, Jan Leike, and Ryan~J. Lowe. 2022.
\newblock \href {https://arxiv.org/abs/2203.02155} {Training language models to follow instructions with human feedback}.
\newblock \emph{ArXiv preprint}, abs/2203.02155.

\bibitem[{Proebsting and Poliak(2024)}]{Proebsting2024HypothesisonlyBI}
Grace Proebsting and Adam Poliak. 2024.
\newblock \href {https://api.semanticscholar.org/CorpusID:273323367} {Hypothesis-only biases in large language model-elicited natural language inference}.
\newblock \emph{ArXiv}, abs/2410.08996.

\bibitem[{Rafailov et~al.(2024)Rafailov, Sharma, Mitchell, Manning, Ermon, and Finn}]{rafailov2024direct}
Rafael Rafailov, Archit Sharma, Eric Mitchell, Christopher~D Manning, Stefano Ermon, and Chelsea Finn. 2024.
\newblock Direct preference optimization: Your language model is secretly a reward model.
\newblock \emph{Advances in Neural Information Processing Systems}, 36.

\bibitem[{Rajbhandari et~al.(2020)Rajbhandari, Rasley, Ruwase, and He}]{rajbhandari2020zero}
Samyam Rajbhandari, Jeff Rasley, Olatunji Ruwase, and Yuxiong He. 2020.
\newblock \href {https://www.microsoft.com/en-us/research/publication/zero-memory-optimizations-toward-training-trillion-parameter-models/} {Zero: Memory optimizations toward training trillion parameter models}.
\newblock ArXiv.

\bibitem[{Riviere et~al.(2024)Riviere, Pathak, Sessa, Hardin, Bhupatiraju, Hussenot, Mesnard, Shahriari, Ram'e, Ferret, Liu, Tafti, Friesen, Casbon, Ramos, Kumar, Lan, Jerome, Tsitsulin, Vieillard, Stańczyk, Girgin, Momchev, Hoffman, Thakoor, Grill, Neyshabur, Walton, Severyn, Parrish, Ahmad, Hutchison, Abdagic, Carl, Shen, Brock, Coenen, Laforge, Paterson, Bastian, Piot, Wu, Royal, Chen, Kumar, Perry, Welty, Choquette-Choo, Sinopalnikov, Weinberger, Vijaykumar, Rogozi'nska, Herbison, Bandy, Wang, Noland, Moreira, Senter, Eltyshev, Visin, Rasskin, Wei, Cameron, Martins, Hashemi, Klimczak-Pluci'nska, Batra, Dhand, Nardini, Mein, Zhou, Svensson, Stanway, Chan, Zhou, Carrasqueira, Iljazi, Becker, Fernandez, van Amersfoort, Gordon, Lipschultz, Newlan, Ji, Mohamed, Badola, Black, Millican, McDonell, Nguyen, Sodhia, Greene, Sjoesund, Usui, Sifre, Heuermann, cia Lago, McNealus, Soares, Kilpatrick, Dixon, Martins, Reid, Singh, Iverson, Gorner, Velloso, Wirth, Davidow, Miller, Rahtz, Watson, Risdal, Kazemi, Moynihan,
  Zhang, Kahng, Park, Rahman, Khatwani, Dao, shad Bardoliwalla, Devanathan, Dumai, Chauhan, Wahltinez, Botarda, Barnes, Barham, Michel, chong Jin, Georgiev, Culliton, Kuppala, Comanescu, Merhej, Jana, Rokni, Agarwal, Mullins, Saadat, Carthy, Perrin, Arnold, bastian Krause, Dai, Garg, Sheth, Ronstrom, Chan, Jordan, Yu, Eccles, Hennigan, Kocisk{\'y}, Doshi, Jain, Yadav, Meshram, Dharmadhikari, Barkley, Wei, Ye, Han, Kwon, Xu, Shen, Gong, Wei, Cotruta, Kirk, Rao, Giang, Peran, Warkentin, Collins, Barral, Ghahramani, Hadsell, Sculley, Banks, Dragan, Petrov, Vinyals, Dean, Hassabis, Kavukcuoglu, Farabet, Buchatskaya, Borgeaud, Fiedel, Joulin, Kenealy, Dadashi, and Andreev}]{Riviere2024Gemma2I}
Gemma Team~Morgane Riviere, Shreya Pathak, Pier~Giuseppe Sessa, Cassidy Hardin, Surya Bhupatiraju, L'eonard Hussenot, Thomas Mesnard, Bobak Shahriari, Alexandre Ram'e, Johan Ferret, Peter Liu, Pouya~Dehghani Tafti, Abe Friesen, Michelle Casbon, Sabela Ramos, Ravin Kumar, Charline~Le Lan, Sammy Jerome, Anton Tsitsulin, and 176 others. 2024.
\newblock \href {https://api.semanticscholar.org/CorpusID:270843326} {Gemma 2: Improving open language models at a practical size}.
\newblock \emph{ArXiv}, abs/2408.00118.

\bibitem[{Stiennon et~al.(2020)Stiennon, Ouyang, Wu, Ziegler, Lowe, Voss, Radford, Amodei, and Christiano}]{Stiennon2020LearningTS}
Nisan Stiennon, Long Ouyang, Jeff Wu, Daniel~M. Ziegler, Ryan~J. Lowe, Chelsea Voss, Alec Radford, Dario Amodei, and Paul Christiano. 2020.
\newblock \href {https://arxiv.org/abs/2009.01325} {Learning to summarize from human feedback}.
\newblock \emph{ArXiv preprint}, abs/2009.01325.

\bibitem[{Wang et~al.(2025{\natexlab{a}})Wang, Gan, Huo, Mu, He, Yang, Li, Xiao, Zhang, Liu et~al.}]{wang2025gram}
Chenglong Wang, Yang Gan, Yifu Huo, Yongyu Mu, Qiaozhi He, Murun Yang, Bei Li, Tong Xiao, Chunliang Zhang, Tongran Liu, and 1 others. 2025{\natexlab{a}}.
\newblock Gram: A generative foundation reward model for reward generalization.
\newblock \emph{arXiv preprint arXiv:2506.14175}.

\bibitem[{Wang et~al.(2025{\natexlab{b}})Wang, Gan, Huo, Mu, Yang, He, Xiao, Zhang, Liu, and Zhu}]{wang2025rovrm}
Chenglong Wang, Yang Gan, Yifu Huo, Yongyu Mu, Murun Yang, Qiaozhi He, Tong Xiao, Chunliang Zhang, Tongran Liu, and Jingbo Zhu. 2025{\natexlab{b}}.
\newblock Rovrm: A robust visual reward model optimized via auxiliary textual preference data.
\newblock In \emph{Proceedings of the AAAI Conference on Artificial Intelligence}, volume~39, pages 25336--25344.

\bibitem[{Wang et~al.(2024{\natexlab{a}})Wang, Zhou, Chang, Li, Mu, Xiao, Liu, and Zhu}]{wang2024hybrid}
Chenglong Wang, Hang Zhou, Kaiyan Chang, Bei Li, Yongyu Mu, Tong Xiao, Tongran Liu, and Jingbo Zhu. 2024{\natexlab{a}}.
\newblock Hybrid alignment training for large language models.
\newblock \emph{arXiv preprint arXiv:2406.15178}.

\bibitem[{Wang et~al.(2024{\natexlab{b}})Wang, Xiong, Xie, Zhao, and Zhang}]{Wang2024InterpretablePV}
Haoxiang Wang, Wei Xiong, Tengyang Xie, Han Zhao, and Tong Zhang. 2024{\natexlab{b}}.
\newblock \href {https://api.semanticscholar.org/CorpusID:270562658} {Interpretable preferences via multi-objective reward modeling and mixture-of-experts}.
\newblock In \emph{Conference on Empirical Methods in Natural Language Processing}.

\bibitem[{Wang et~al.(2024{\natexlab{c}})Wang, Bukharin, Delalleau, Egert, Shen, Zeng, Kuchaiev, and Dong}]{Wang2024HelpSteer2PreferenceCR}
Zhilin Wang, Alexander Bukharin, Olivier Delalleau, Daniel Egert, Gerald Shen, Jiaqi Zeng, Oleksii Kuchaiev, and Yi~Dong. 2024{\natexlab{c}}.
\newblock \href {https://api.semanticscholar.org/CorpusID:273025954} {Helpsteer2-preference: Complementing ratings with preferences}.
\newblock \emph{ArXiv}, abs/2410.01257.

\bibitem[{Xiao and Zhu(2025)}]{xiao2025foundations}
Tong Xiao and Jingbo Zhu. 2025.
\newblock Foundations of large language models.
\newblock \emph{arXiv preprint arXiv:2501.09223}.

\bibitem[{Yang et~al.(2025)Yang, Li, Yang, Zhang, Hui, Zheng, Yu, Gao, Huang, Lv, Zheng, Liu, Zhou, Huang, Hu, Ge, Wei, Lin, Tang, Yang, Tu, Zhang, Yang, Yang, Zhou, Zhou, Lin, Dang, Bao, Yang, Yu, Deng, Li, Xue, Li, Zhang, Wang, Zhu, Men, Gao, Liu, Luo, Li, Tang, Yin, Ren, Wang, Zhang, Ren, Fan, Su, Zhang, Zhang, Wan, Liu, Wang, Cui, Zhang, Zhou, and Qiu}]{Yang2025Qwen3TR}
An~Yang, Anfeng Li, Baosong Yang, Beichen Zhang, Binyuan Hui, Bo~Zheng, Bowen Yu, Chang Gao, Chengen Huang, Chenxu Lv, Chujie Zheng, Dayiheng Liu, Fan Zhou, Fei Huang, Feng Hu, Hao Ge, Haoran Wei, Huan Lin, Jialong Tang, and 41 others. 2025.
\newblock \href {https://api.semanticscholar.org/CorpusID:278602855} {Qwen3 technical report}.
\newblock \emph{ArXiv}, abs/2505.09388.

\bibitem[{Yang et~al.(2024)Yang, Ding, Lin, Zhang, and Zhang}]{Yang2024RegularizingHS}
Rui Yang, Ruomeng Ding, Yong Lin, Huan Zhang, and Tong Zhang. 2024.
\newblock \href {https://api.semanticscholar.org/CorpusID:270521260} {Regularizing hidden states enables learning generalizable reward model for llms}.
\newblock \emph{ArXiv}, abs/2406.10216.

\bibitem[{Zheng et~al.(2024)Zheng, Zhang, Zhang, Ye, Luo, and Ma}]{Zheng2024LlamaFactoryUE}
Yaowei Zheng, Richong Zhang, Junhao Zhang, Yanhan Ye, Zheyan Luo, and Yongqiang Ma. 2024.
\newblock \href {https://api.semanticscholar.org/CorpusID:268536974} {Llamafactory: Unified efficient fine-tuning of 100+ language models}.
\newblock \emph{ArXiv}, abs/2403.13372.

\bibitem[{Zhou et~al.(2023)Zhou, Liu, Xu, Iyer, Sun, Mao, Ma, Efrat, Yu, Yu, Zhang, Ghosh, Lewis, Zettlemoyer, and Levy}]{Zhou2023LIMALI}
Chunting Zhou, Pengfei Liu, Puxin Xu, Srini Iyer, Jiao Sun, Yuning Mao, Xuezhe Ma, Avia Efrat, Ping Yu, L.~Yu, Susan Zhang, Gargi Ghosh, Mike Lewis, Luke Zettlemoyer, and Omer Levy. 2023.
\newblock \href {https://api.semanticscholar.org/CorpusID:258822910} {Lima: Less is more for alignment}.
\newblock \emph{ArXiv}, abs/2305.11206.

\bibitem[{Zhou et~al.(2024)Zhou, Wang, Hu, Xiao, Zhang, and Zhu}]{zhou2024prior}
Hang Zhou, Chenglong Wang, Yimin Hu, Tong Xiao, Chunliang Zhang, and Jingbo Zhu. 2024.
\newblock Prior constraints-based reward model training for aligning large language models.
\newblock In \emph{China National Conference on Chinese Computational Linguistics}, pages 555--570. Springer.

\end{thebibliography}

\clearpage

\appendix

\section{Implementation Details}

\subsection{Construction of UniHypoBench}
\label{app:construct-unihypo}
The \textit{hypothesis-driven} nature of our approach necessitates careful attention to preserving diversity across hypotheses, motivating us to construct UniHypoBench, a benchmark maintaining 10 or more hypotheses per input. 
We construct UniHypoBench based on the RewardBench instruction set, leveraging its comprehensive coverage of diverse task types.
Our benchmark construction process begins by collecting hypothesis samples from multiple powerful commercial and open-source LLMs, as specified in Table \ref{tab:applied-models}.
\begin{table}[t]
    \centering
    \scalebox{0.81}{
    \begin{tabular}{lcc}
\toprule[1.1pt]
Name & Commercial & Open-source \\ \cmidrule(r){1-3}
Claude-3-Haiku & \checkmark & \ding{55} \\
GPT-4o & \checkmark & \ding{55} \\
DeepSeek-V2-Lite & \ding{55} & \checkmark \\
Qwen2.5-32B & \ding{55} & \checkmark \\
Qwen-14B & \ding{55} & \checkmark \\
Mixtral-8x7B-Instruct-v0.1& \ding{55} & \checkmark \\
LLaMA-3-8B-Instruct & \ding{55} & \checkmark \\
ChatGLM3-6B & \ding{55} & \checkmark \\
\bottomrule[1.1pt]
\end{tabular}
    }
    \caption{Models selected for UniHypoBench construction.}
    \label{tab:applied-models}
\end{table}

The UniHypoBench is constructed considering a balance between the response diversity and overall quality.
Therefore, the construction involves two key components: 
(1) \textit{Diverse Model Selection:} Our model selection encompasses diverse LLM architectures, scales, and training paradigms to maximize hypothesis variation, as highlighted in Table \ref{tab:applied-models}; 
(2) \textit{Increased Randomness during Sampling:} We conduct sampling with a temperature of 0.75 and top-p sampling at 0.95 that carefully balance diversity and output quality.

To enhance the response diversity while maintaining quality, we configured the sampling parameters with a temperature setting of 0.75 and top-p value of 0.95, with all responses truncated at 768 tokens.
Following generation, we implemented a filtering process to remove low-quality and empty responses, thereby ensuring the benchmark's reliability and consistency.

The Table \ref{tab:data-example} shows an example of the UniHypoBench, and we have also created a repository for the dataset: \href{https://anonymous.4open.science/r/HEAL-4C68/}{https://anonymous.4open.science/r/HEAL-4C68/}


\subsection{Experimental Setups}
\label{app:training-setup}


Our implementation leverages \texttt{LLaMA-Factory} \cite{Zheng2024LlamaFactoryUE} for model training and \texttt{vLLM} \cite{Kwon2023EfficientMM} for efficient inference.
All experiments were conducted on 2$\times$NVIDIA 3090 GPUs, with additional optimization through \texttt{DeepSpeed} \cite{rajbhandari2020zero} ZeRO-2 to minimize computational overhead and accelerate training.
Following established practices in preference optimization \cite{Meng2024SimPOSP}, we maintain an effective batch size of 128 and employed a cosine learning rate schedule with 10\% warmup steps.
To balance computational efficiency with model performance, we set the training sequence length to 1024 tokens.

Before final model training, we performed extensive hyperparameter tuning to identify optimal configurations for each method.
We first search the learning rates invidually in the range of [3e-7, 7e-7, 1e-6].
Then we search method-specific parameters whose search ranges are detailed in Table \ref{tab:hyperparameter-info}.

\begin{table*}[t]
    \centering


\begin{tabular}{lll}
\toprule[1.1pt]
Method & Objective & Hyperparameter \\ \cmidrule(r){1-3}
\multirow{2}{*}{DPO} & \multirow{2}{*}{$-\log\sigma\left(\beta\log\frac{\pi_\theta(y_w|x)}{\pi_\mathrm{ref}(y_w|x)}-\beta\log\frac{\pi_\theta(y_l|x)}{\pi_\mathrm{ref}(y_l|x)}\right)$} & \multirow{2}{*}{ $\beta \in [0.01, 0.05, 0.1]$} \\
 &  &  \\ \cmidrule(r){1-3}
\multirow{2}{*}{ORPO} & $-\log p_\theta(y_w|x)-\lambda\log\sigma\left(\log\frac{p_\theta(y_w|x)}{1-p_\theta(y_w|x)}-\log\frac{p_\theta(y_l|x)}{1-p_\theta(y_l|x)}\right),$& \multirow{2}{*}{$\lambda \in [0.1, 0.5, 1.0]$} \\
 & where $p_\theta(y|x)=\mathrm{exp}\left(\frac{1}{|y|}\log\pi_\theta(y|x)\right)$ &  \\ \cmidrule(r){1-3}
\multirow{2}{*}{SimPO} & \multirow{2}{*}{$-\log\sigma\left(\frac{\beta}{|y_w|}\log\pi_\theta(y_w|x)-\frac{\beta}{|y_l|}\log\pi_\theta(y_l|x)-\gamma\right)$} & $\beta \in [2.0, 2.5, 3.0, 5.0, 10.0],$ \\
 &  &$ \gamma \in [0.3, 0.5, 1.0]$ \\

\bottomrule[1.1pt]
\end{tabular}
    \caption{Optimization objectives and hyperparameter search ranges of applied preference learning methods}
    \label{tab:hyperparameter-info}
\end{table*}

\subsection{Guidelines on Incorporating Personalized Preferences}
In this paper, we define a reference preference as a \textit{gold-standard} that serves as the optimization target for LLM alignment. 
It is important to note that this reference is not a universal ground truth; rather, it is derived from a specific preference distribution, such as one embodied by a reward model or a particular group of human annotators. 
While the inherent diversity of human values makes a single, universally valid \textit{gold-standard} infeasible, HEAL can effectively estimate user-specific preferences by leveraging ranking-based annotations that capture distinct preference patterns.
HEAL is designed for straightforward adaptation to personalized preferences. 
The process for applying it to diverse user groups involves the following key steps:
\begin{itemize}
    \item \textbf{Data Generation:} Construct a set of 1-to-n input-response pairs. This can be achieved by sampling from one or more LLMs or by using pre-existing datasets.
    \item \textbf{Preference Annotation:} The generated responses for each input must be ranked. This can be performed using one of two methods:
        \begin{itemize}
            \item Engaging human annotators who are representative of the target user group's preferences.
            \item Employing a reward model fine-tuned on the desired preference distribution.
        \end{itemize}
    A critical requirement for this stage is that the annotators or proxy models must accurately reflect the target preferences.
    \item \textbf{Applying HEAL for Alignment Quantification:} The resulting ranked list from the annotation step establishes the \textit{gold-standard} for that specific user group. HEAL can then be utilized to quantify the alignment of any generative model, reward model, or annotator with this bespoke reference.
\end{itemize}

\section{Results on Other Backbone Models}

To further validate the generalizability of our framework, we extend the main experiment to incorporate four additional backbone models: Gemma \cite{Riviere2024Gemma2I}, Mistral \cite{jiang2023clip}, LLaMA-3.1 \cite{Dubey2024TheL3}, and Qwen3 \cite{Yang2025Qwen3TR}. 
The comprehensive performance comparison across all evaluated models is presented in Table \ref{tab:main-result-append}.
These results consistently reinforce our paper's central finding that preference optimization effectively captures preference information. 
While the statistical significance of these metrics is slightly lower than our primary results—a difference we attribute to the absence of model-specific hyperparameter tuning—the overall trend remains robust and aligns with our main conclusion.

\begin{table*}[t]
    \centering
    \scalebox{0.76}{
\begin{tabular}{lrrrrrrrrrrrr}
\toprule[1.1pt]
\multicolumn{1}{c}{\multirow{3}{*}{\textbf{\begin{tabular}[c]{@{}c@{}}Model/Method\end{tabular}}}} & \multicolumn{6}{c}{\textbf{w/o Length Normalization}} & \multicolumn{6}{c}{\textbf{w/ Length Normalization}} \\ \cmidrule(r){2-7} \cmidrule(r){8-13}
\multicolumn{1}{c}{} & \multicolumn{2}{c}{\textbf{UniHypo}} & \multicolumn{2}{c}{\textbf{HelpSteer2}} & \multicolumn{2}{c}{\textbf{UltraFeedback}} & \multicolumn{2}{c}{\textbf{UniHypo}} & \multicolumn{2}{c}{\textbf{HelpSteer2}} & \multicolumn{2}{c}{\textbf{UltraFeedback}} \\ \cmidrule(r){2-3} \cmidrule(r){4-5} \cmidrule(r){6-7} \cmidrule(r){8-9} \cmidrule(r){10-11} \cmidrule(r){12-13}    
\multicolumn{1}{c}{} & \multicolumn{1}{r}{\textbf{RA}} & \multicolumn{1}{r}{\textbf{PSC}} & \multicolumn{1}{r}{\textbf{RA}} & \multicolumn{1}{r}{\textbf{PSC}} & \multicolumn{1}{r}{\textbf{RA}} & \multicolumn{1}{r}{\textbf{PSC}} & \multicolumn{1}{r}{\textbf{RA}} & \multicolumn{1}{r}{\textbf{PSC}} & \multicolumn{1}{r}{\textbf{RA}} & \multicolumn{1}{r}{\textbf{PSC}} & \multicolumn{1}{r}{\textbf{RA}} & \multicolumn{1}{r}{\textbf{PSC}} \\ 
\cmidrule(r){1-13}
\multicolumn{13}{l}{\textit{\textbf{Alignment with ArmoRM-Llama3-8B-v0.1 (Same Preference Distribution)}}} \\
\cmidrule(r){1-13}
Gemma-2-9B                      & \textbf{53.21} &  \textbf{0.128} & \textbf{48.16} & \textbf{-0.037} & 54.06 & 0.103 & 47.33 & \textbf{-0.047} & \textbf{53.21} &  \textbf{0.064} & 47.74 & -0.060 \\
\qquad\qquad\qquad +DPO         & 52.04 &  0.096 & 47.11 & -0.058 & \textbf{54.15} & \textbf{0.105} & 47.10 & -0.058 & 52.31 &  0.046 & \underline{48.17} & \textbf{-0.050} \\
\qquad\qquad\qquad +ORPO        & \underline{52.10} &  \underline{0.098} & 46.90 & -0.062 & 53.91 & 0.100 & \textbf{47.18} & \underline{-0.056} & \underline{52.41} &  \underline{0.048} & 48.07 & -0.053 \\ 
\qquad\qquad\qquad +SimPO       & 52.07 &  0.097 & \underline{47.36} & \underline{-0.053} & \underline{54.09} & \underline{0.104} & \underline{47.17} & -0.057 & 51.73 &  0.035 & \textbf{48.21} & \underline{-0.051} \\ \cmidrule(r){1-13}
Mistral-7B-it-v0.3              & 48.11 & -0.006 & \textbf{45.54} & \textbf{-0.089} & 51.20 & \underline{0.035} & 42.63 & \underline{-0.155} & 46.90 & \underline{-0.062} & 46.47 & \underline{-0.081} \\
\qquad\qquad\qquad +DPO         & \underline{48.13} & \underline{-0.005} & \underline{45.53} & \textbf{-0.089} & 51.21 & \underline{0.035} & \underline{42.66} & \underline{-0.155} & \underline{46.91} & \underline{-0.062} & 46.41 & \underline{-0.081} \\
\qquad\qquad\qquad +ORPO        & \textbf{48.24} & \textbf{-0.003} & 45.52 & \underline{-0.090} & \textbf{51.36} & \textbf{0.036} & \textbf{42.84} & \textbf{-0.149} & \textbf{47.14} & \textbf{-0.057} & \textbf{46.64} & \textbf{-0.078} \\
\qquad\qquad\qquad +SimPO       & \underline{48.13} & \underline{-0.005} & \textbf{45.54} & \textbf{-0.089} & \underline{51.24} & 0.034 & 42.65 & \underline{-0.155} & 46.90 & \underline{-0.062} & \underline{46.54} & \underline{-0.081} \\ \cmidrule(r){1-13}
LLaMA-3.1-8B-Instruct           & \underline{55.62} &  \textbf{0.172} & 47.71 & \underline{-0.046} & \textbf{54.05} & \underline{0.099} & 48.72 & \textbf{-0.040} & \textbf{52.41} &  \textbf{0.048} & \underline{50.34} &  \underline{0.011} \\
\qquad\qquad\qquad +DPO         & \textbf{55.63} &  \textbf{0.172} & \textbf{47.94} & \textbf{-0.041} & \underline{53.98} & \textbf{0.100} & \underline{48.74} & \textbf{-0.040} & \underline{52.30} &  \underline{0.046} & 50.28 &  \underline{0.011} \\
\qquad\qquad\qquad +ORPO        & 55.61 &  \textbf{0.172} & \underline{47.93} & \textbf{-0.041} & 53.84 & 0.098 & 48.72 & \textbf{-0.040} & 52.19 &  0.044 & 50.27 &  0.010 \\ 
\qquad\qquad\qquad +SimPO       & \underline{55.62} &  \textbf{0.172} & 47.71 & \underline{-0.046} & 53.97 & \underline{0.099} & \textbf{48.75} & \textbf{-0.040} & \textbf{52.41} &  \textbf{0.048} & \textbf{50.35} &  \textbf{0.012} \\ \cmidrule(r){1-13}
Qwen3-8B                        & \underline{51.12} &  \textbf{0.064} & 45.85 & -0.083 & \underline{51.99} & \underline{0.053} & 46.77 & \underline{-0.074} & \textbf{48.97} & \textbf{-0.021} & \underline{46.07} & \underline{-0.093} \\
\qquad\qquad\qquad +DPO         & \underline{51.12} &  \textbf{0.064} & \textbf{51.13} &  \textbf{0.064} & 51.96 & \textbf{0.054} & \underline{46.79} & \textbf{-0.073} & 46.77 & -0.074 & \underline{46.07} & \textbf{-0.092} \\
\qquad\qquad\qquad +ORPO        & \textbf{51.14} &  \textbf{0.064} & 45.75 & -0.085 & \underline{51.99} & \underline{0.053} & \textbf{46.81} & \textbf{-0.073} & 48.62 & -0.028 & \textbf{46.27} & \textbf{-0.092} \\
\qquad\qquad\qquad +SimPO       & \underline{51.12} &  \textbf{0.064} & \underline{45.98} & \underline{-0.080} & \textbf{52.03} & \textbf{0.054} & 46.78 & \textbf{-0.073} & \underline{48.85} & \underline{-0.023} & 46.06 & \textbf{-0.092} \\
\bottomrule[1.1pt]
\end{tabular}}
    \caption{
        Additional experimental results on different preference optimization methods.
        RA and PSC denote ranking accuracy and preference strength correlation, respectively.
        The best results for each group are in \textbf{bold}.
        The second-best results for each group are with \underline{underline}.
    }
    \label{tab:main-result-append}
\end{table*}

\section{More Analysis}
\label{sec:more-analysis}
\subsection{Preference Learning Achieves Limited Improvements with Confident LLMs.}  

Building upon the main results presented in Table \ref{tab:main-result}, we observe that preference optimization yields limited improvement for LLaMA-3.2-3B-Instruct, with both ranking accuracy and preference strength showing marginal gains or even performance degradation. 
This unexpected outcome suggests potential overfitting to the original training corpus during earlier optimization stages. 
More fundamentally, these findings reveal an important relationship between a base model's core capabilities and its capacity for effective preference learning - implying that successful alignment may be constrained by the underlying model's basic capabilities before fine-tuning.




\subsection{Relationship with Existing Methods.}
\label{sec:relation-exist-work}
HEAL's methodology for assessing alignment is fundamentally different from conventional techniques, as outlined in Section \ref{sec:discussion}. 
To empirically validate this distinction, we benchmarked HEAL against leading metrics—pairwise accuracy (RewardBench) and win-rate (Alpaca-Eval \cite{Dubois2024LengthControlledAA}, Arena-Hard \cite{Li2024FromCD} )—utilizing data from \cite{Meng2024SimPOSP} (Table \ref{tab:exist-work}).

Our analysis reveals a moderate correlation between HEAL and pairwise accuracy methods, confirming its unique methodological foundation. 
This finding strongly suggests that HEAL captures facets of alignment that are not fully addressed by traditional metrics. 
Beyond its novel measurement capabilities, HEAL's \textit{sampling-free} architecture presents a computationally efficient alternative to expensive \textit{LLM-as-a-Judge} protocols.

We have also listed several key advantages of HEAL in Table \ref{tab:key-advantages}.
Most notably, HEAL advances beyond prior approaches by introducing a unified \textit{hypothesis-driven} framework that seamlessly integrates generative models, reward models, and human evaluation. 
This distinctive architecture offers flexibility in alignment measurement—a core contribution of our work. Below we highlight HEAL’s key advantages:
\begin{itemize}
    \item \textbf{Unification:} HEAL provides a \textit{hypothesis-based} evaluation paradigm applicable to both generative and reward models, enabling direct comparison across different model types.
    \item \textbf{Scalability:} HEAL flexibly adapts to diverse preference distributions and model architectures without requiring structural modifications.
    \item \textbf{Consistency:} As a \textit{sampling-free} method, HEAL yields deterministic results, eliminating the need for repeated testing to optimize hyperparameters.
    \item \textbf{Cost-Efficiency:} Unlike \textit{LLM-as-a-Judge} approaches (\textit{e.g.}, Alpaca-Eval), HEAL operates without external API calls, significantly reducing evaluation costs.
\end{itemize}

\begin{table*}[t]
    \centering
    \scalebox{0.90}{

\begin{tabular}{lcccccc}
\toprule[1.1pt]
\multirow{2}{*}{Model/Method} & \multicolumn{2}{c}{HEAL} & RewardBench & \multicolumn{2}{l}{Alpaca-Eval} & Arena-Hard \\ 
 & RA. & PSC. & ACC. & LC. & WR. & WR. \\ \cmidrule(r){1-7}
LLaMA-3-8B-Instruct & 54.15 & 0.124 & 72.66 & 26.0 & 25.3 & 22.3 \\
+DPO & 54.16 & 0.138 & 72.63 & 48.2 & 47.5 & 35.2 \\
+SimPO & 63.59 & 0.502 & 72.70 & 53.7 & 47.5 & 36.5 \\
+ORPO & 52.67 & 0.084 & 72.23 & 38.1 & 33.8 & 28.2 \\
\bottomrule[1.1pt]
\end{tabular}
    }
    \caption{Performance Metrics for Alignment Evaluation Methods.}
    \label{tab:exist-work}
\end{table*}

\begin{table*}[t]
    \centering
    \scalebox{0.90}{

\begin{tabular}{lccc}
\toprule[1.1pt]
Feature & RewardBench & \textit{LLM-as-a-Judge} & HEAL \\ \cmidrule(r){1-4}
Reward Model Evaluation & \checkmark & \ding{55} & \checkmark \\
Generative Model Evaluation & \ding{55} & \checkmark & \checkmark \\
Scalability & \checkmark & \checkmark & \checkmark \\
Reference Required & \checkmark & \checkmark & \checkmark \\
Hyper-parameter Free & \checkmark & \ding{55} & \checkmark \\
API Free & \checkmark & \ding{55} & \checkmark \\
Sampling Free & \checkmark & \ding{55} & \checkmark \\
\bottomrule[1.1pt]
\end{tabular}
    }
    \caption{Key advantages of HEAL compared with existing methods.}
    \label{tab:key-advantages}
\end{table*}

\section{More Upset Plots}

\begin{figure*}
    \centering
    \subfigure[UniHypo (w/o Length Normalized)]{
        \scalebox{1.5}{\includegraphics[scale=0.43]{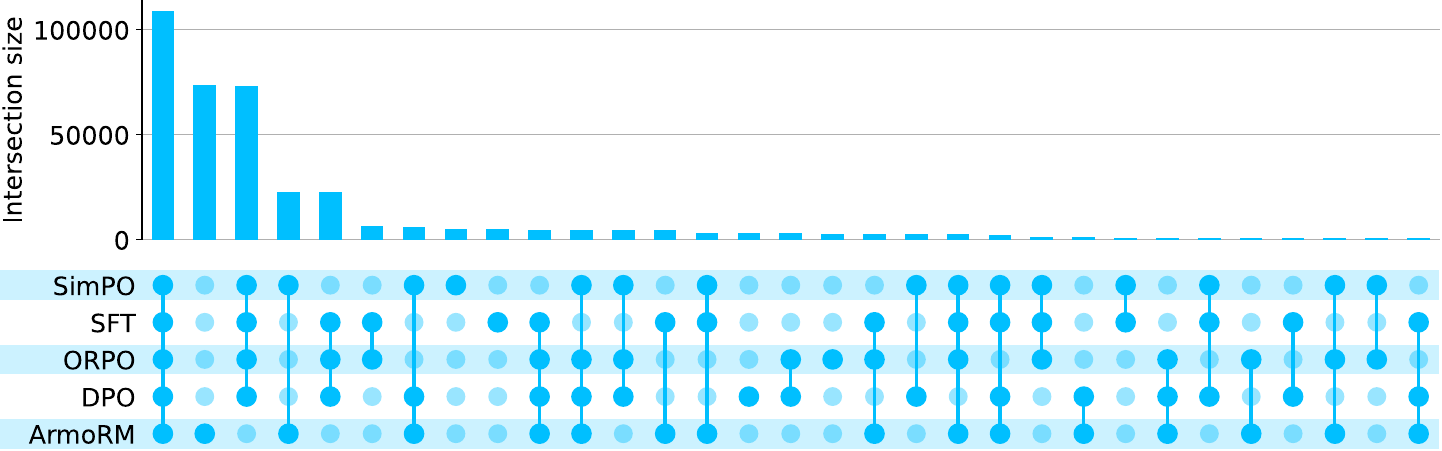}}
    }
    \subfigure[HelpSteer2 (w/o Length Normalized)]{
        \scalebox{1.5}{\includegraphics[scale=0.43]{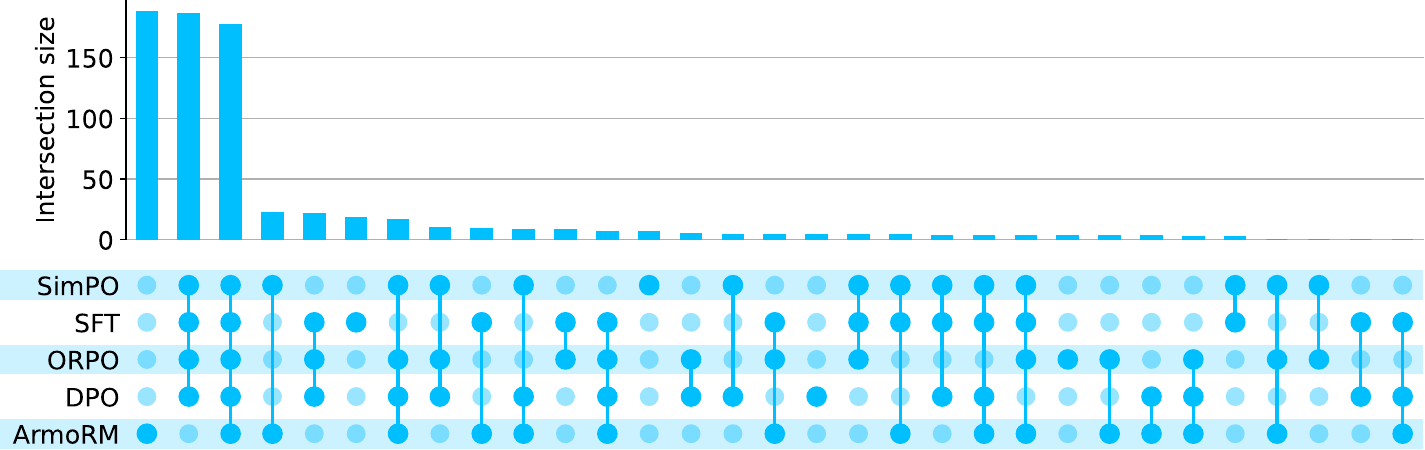}}
    }
    \subfigure[UltraFeedback (w/o Length Normalized)]{
        \scalebox{1.5}{\includegraphics[scale=0.43]{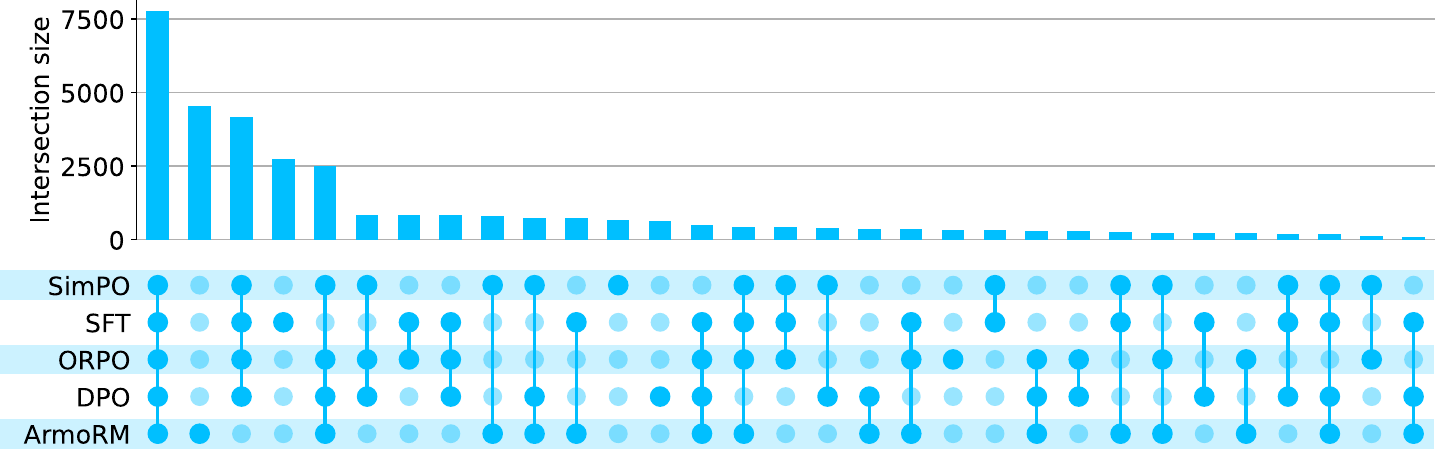}}
    }
    \vspace{-0.3cm}
    \caption{
        Upset plots of generation likelihoods without length normalization.
    }
    \vspace{-0.3cm}
    \label{fig:upset-plots-wo-ln}
\end{figure*}

\begin{figure*}
    \centering
    \subfigure[UniHypo (w/ Length Normalized)]{
        \scalebox{1.5}{\includegraphics[scale=0.43]{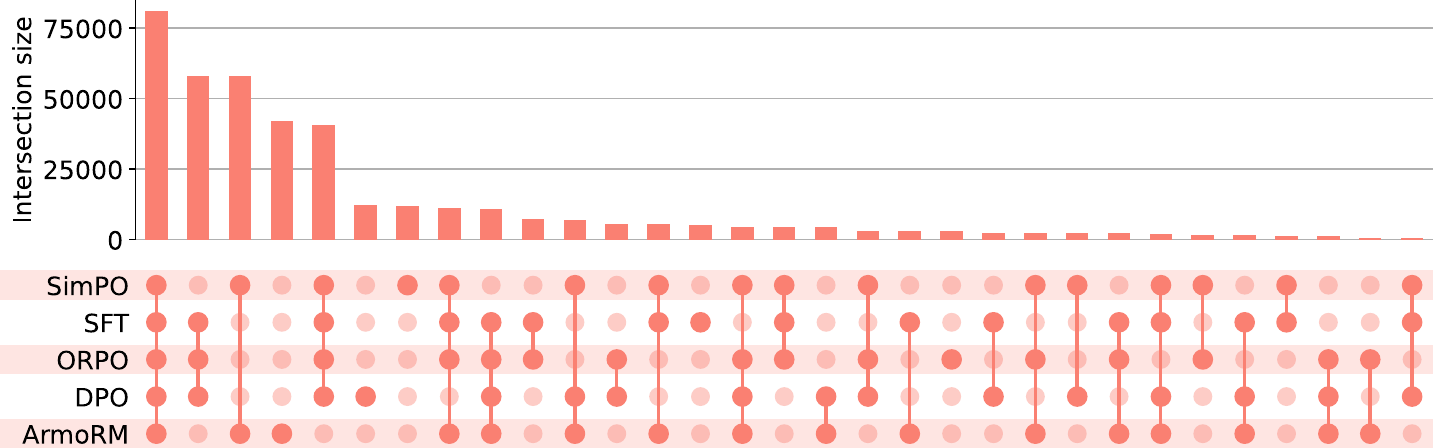}}
    }
    \subfigure[HelpSteer2 (w/ Length Normalized)]{
        \scalebox{1.5}{\includegraphics[scale=0.43]{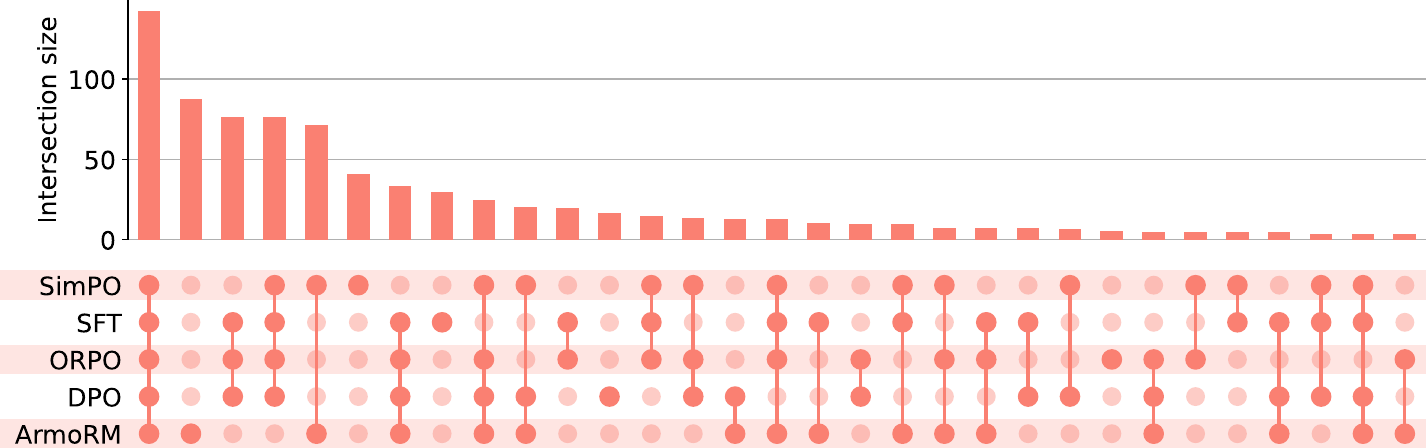}}
    }
    \subfigure[UltraFeedback (w/ Length Normalized)]{
        \scalebox{1.5}{\includegraphics[scale=0.43]{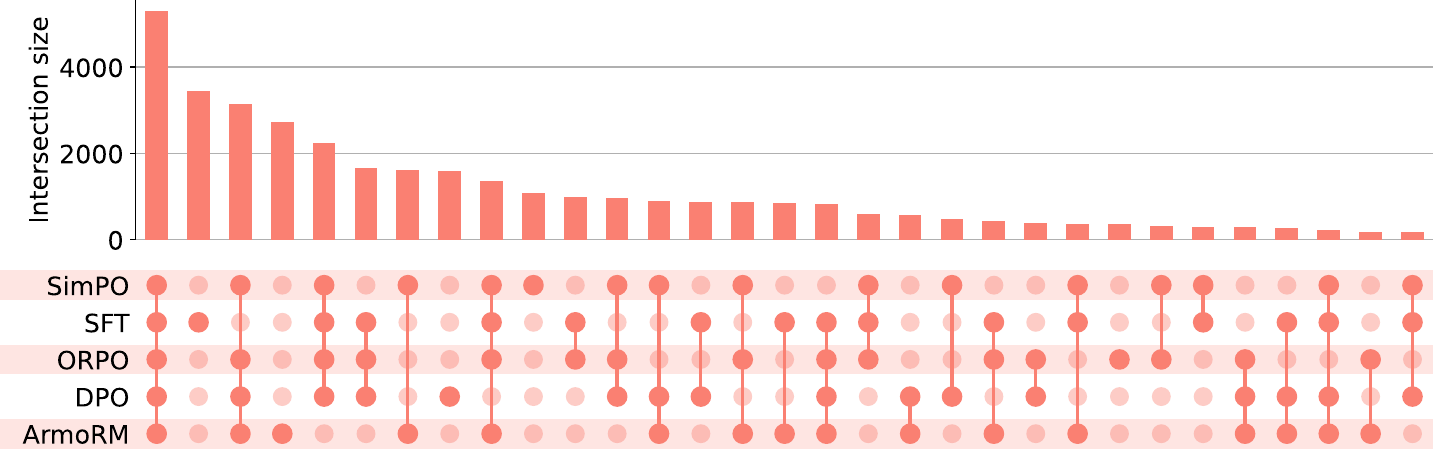}}
    }
    \vspace{-0.3cm}
    \caption{
        Upset plots of generation likelihoods with length normalization.
    }
    \vspace{-0.3cm}
    \label{fig:upset-plots-w-ln}
\end{figure*}

This section presents the complete upset plot visualizations in Figure \ref{fig:upset-plots-wo-ln}, along with their length-normalized counterparts in Figure \ref{fig:upset-plots-w-ln}. 
The observed patterns remain consistent with our preliminary analysis in Figure \ref{fig:upset-plot}, further validating our earlier conclusions regarding preference alignment behaviors.
Notably, we find that the length normalization has smoothed the distribution of the intersections, which could be valuable for further study.


\section{Joint Distribution of Reward Scores and Generation Likelihoods}
\begin{figure*}
    \centering
    \subfigure[UniHypo - SFT]{
        \scalebox{0.75}{\includegraphics[scale=0.43]{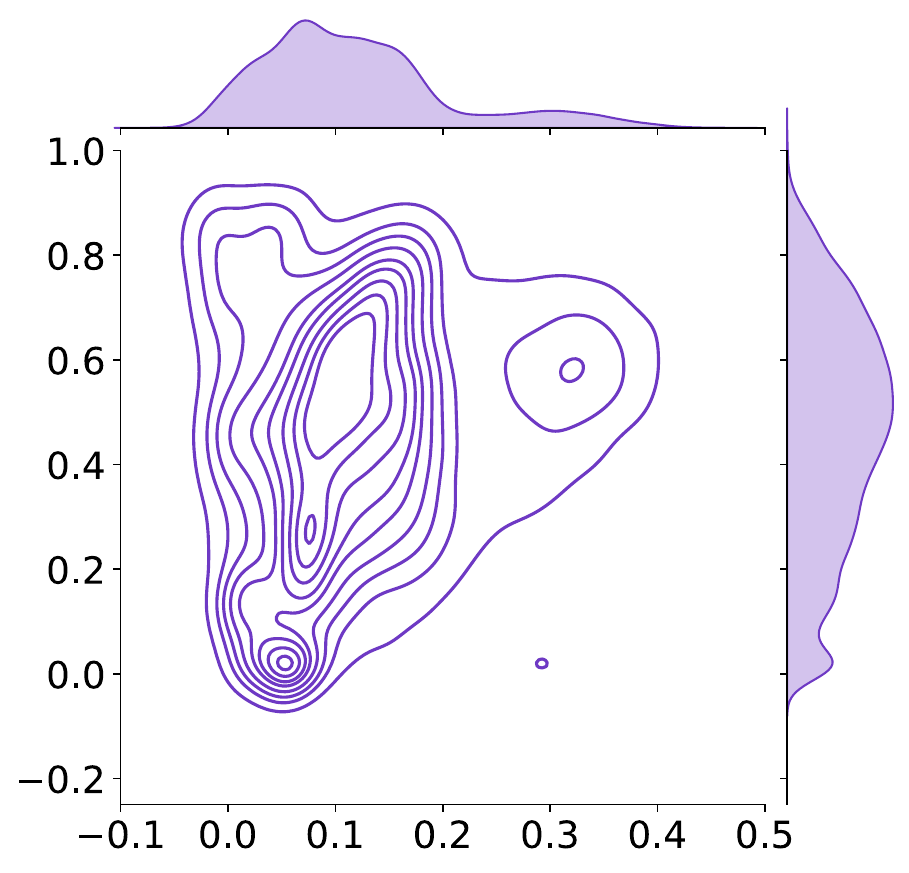}}
    }
    \subfigure[HelpSteer2 - SFT]{
        \scalebox{0.75}{\includegraphics[scale=0.43]{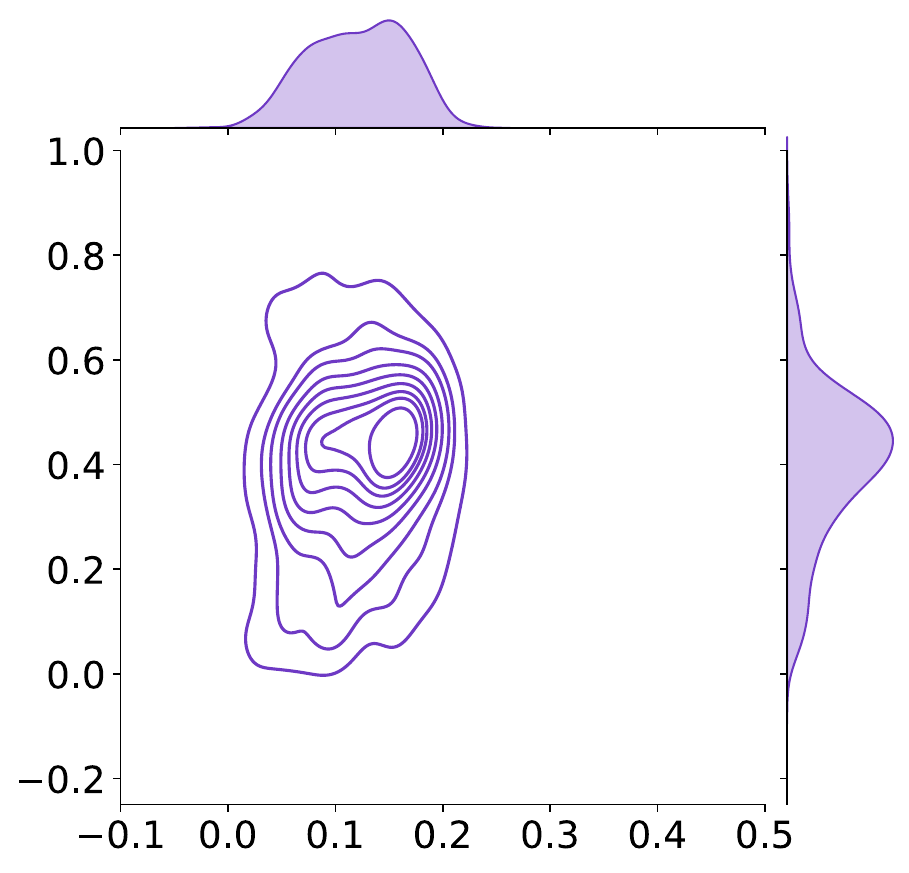}}
    }
    \subfigure[UltraFeedback - SFT]{
        \scalebox{0.75}{\includegraphics[scale=0.43]{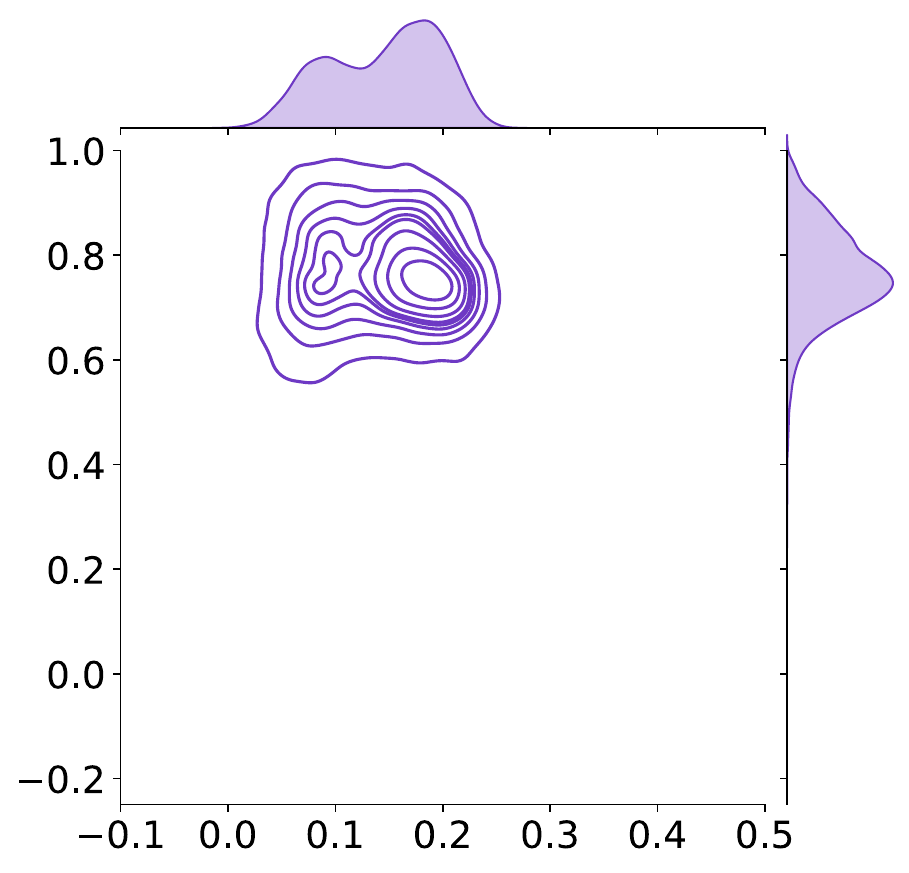}}
    }    
    \subfigure[UniHypo - DPO]{
        \scalebox{0.75}{\includegraphics[scale=0.43]{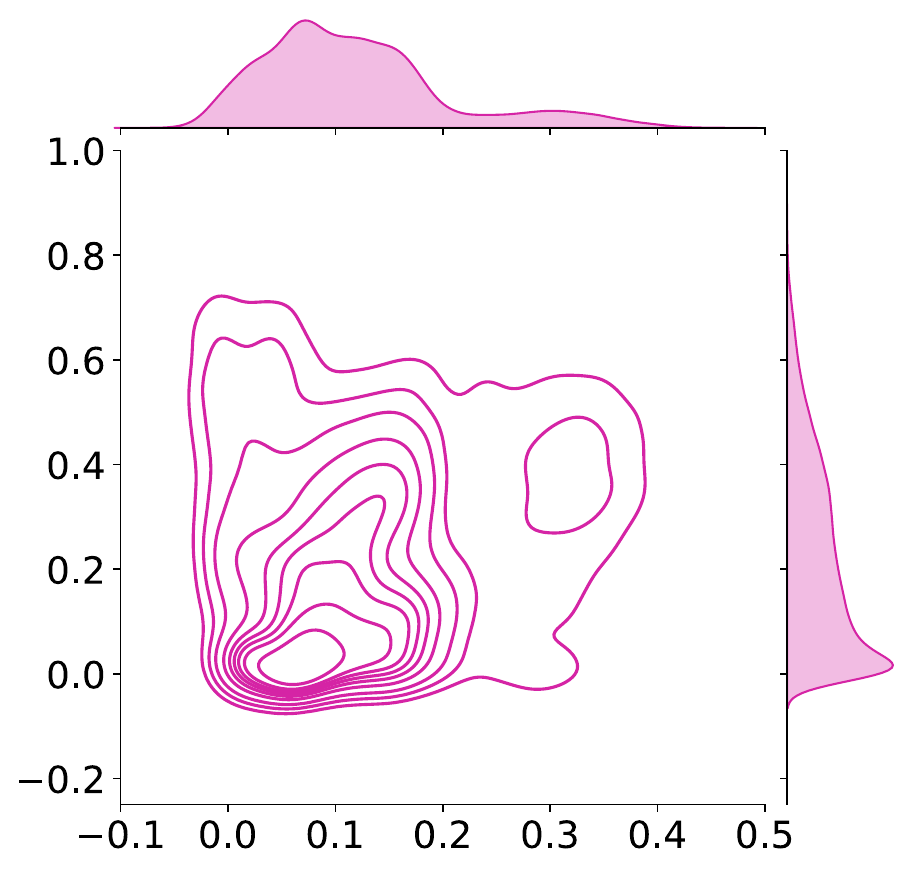}}
    }
    \subfigure[HelpSteer2 - DPO]{
        \scalebox{0.75}{\includegraphics[scale=0.43]{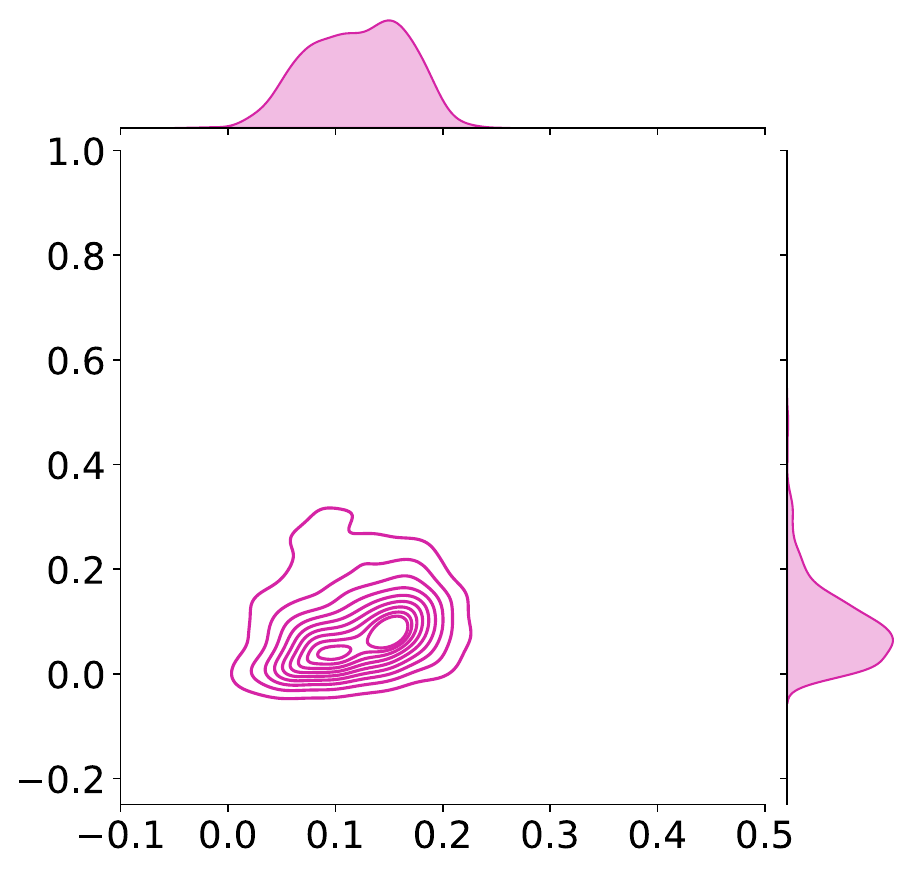}}
    }
    \subfigure[UltraFeedback - DPO]{
        \scalebox{0.75}{\includegraphics[scale=0.43]{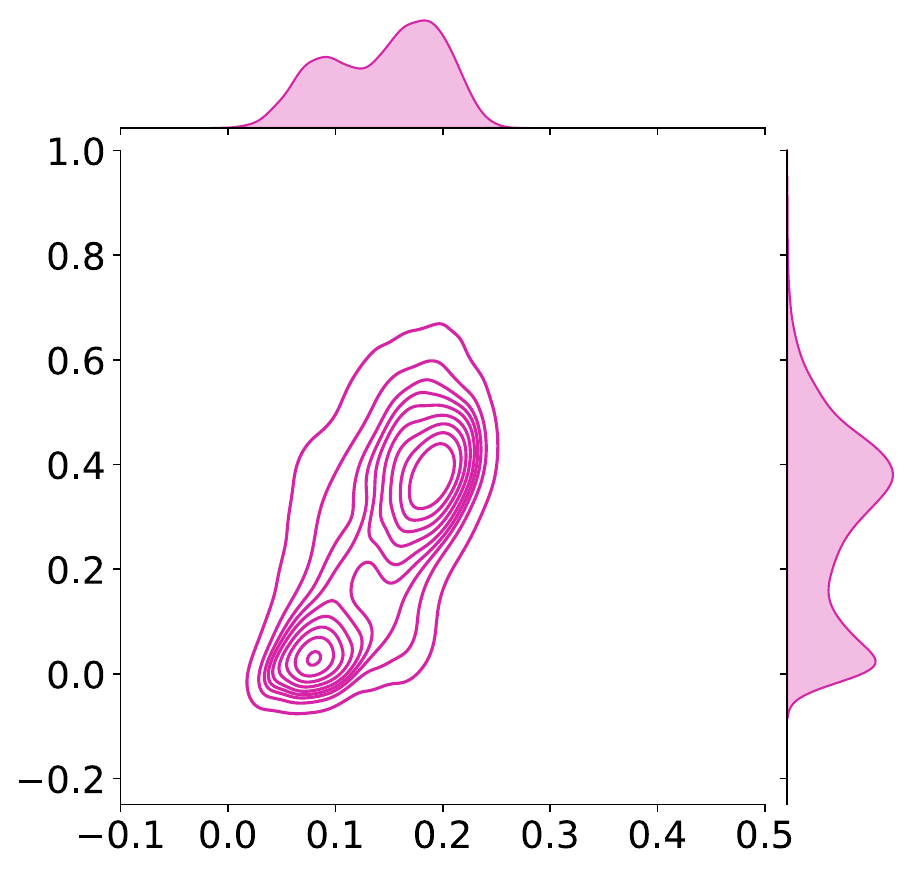}}
    }    
    \subfigure[UniHypo - ORPO]{
        \scalebox{0.75}{\includegraphics[scale=0.43]{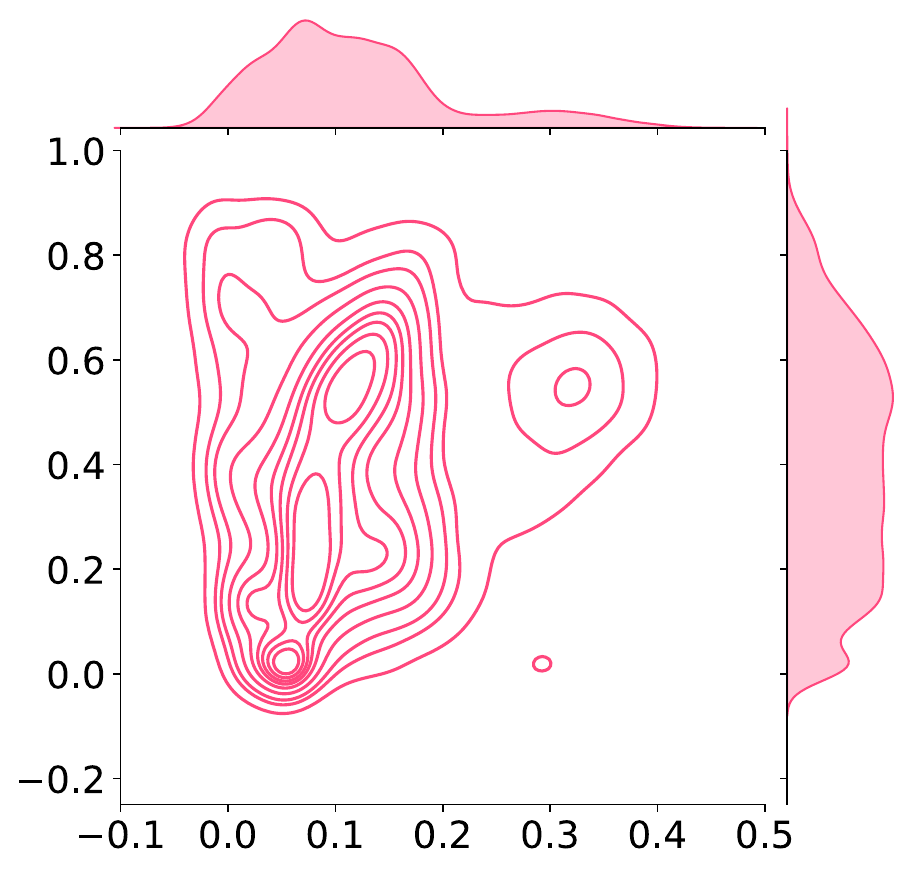}}
    }
    \subfigure[HelpSteer2 - ORPO]{
        \scalebox{0.75}{\includegraphics[scale=0.43]{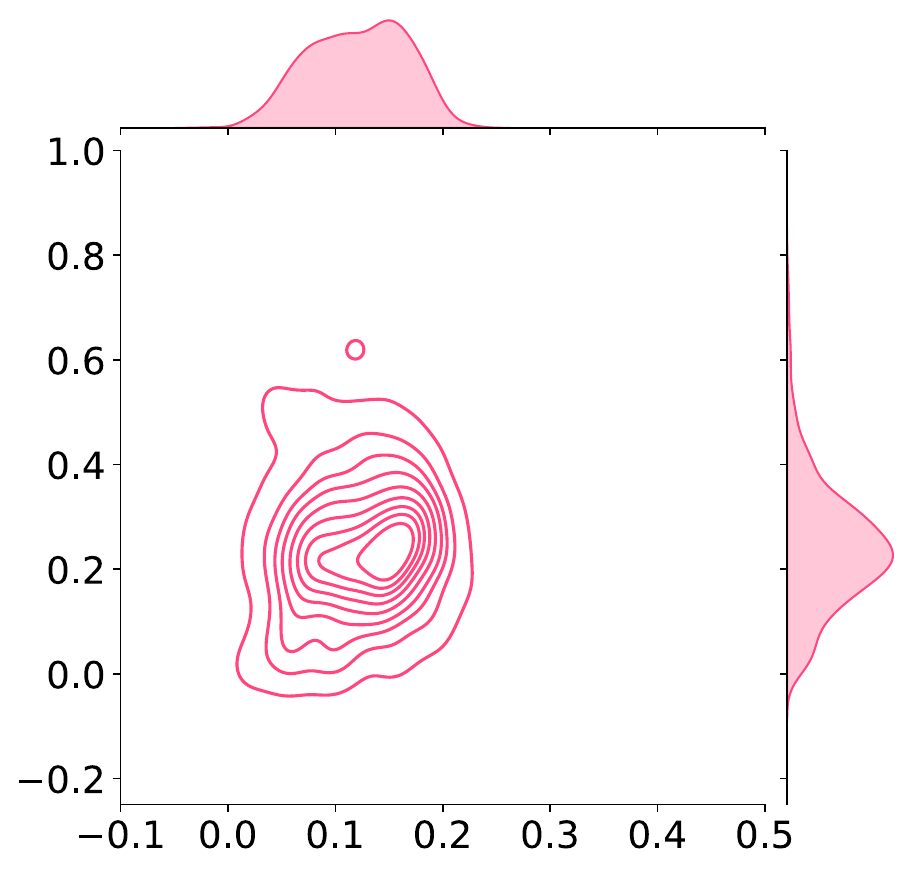}}
    } 
    \subfigure[UltraFeedback - ORPO]{
        \scalebox{0.75}{\includegraphics[scale=0.43]{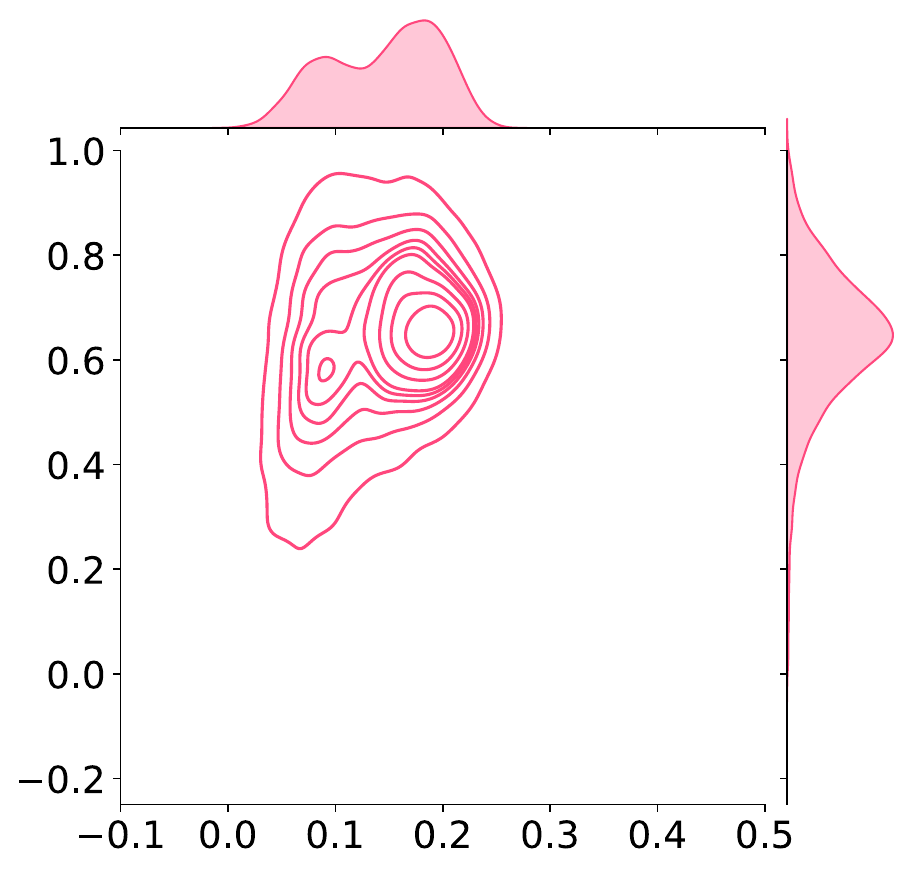}}
    }    
    \subfigure[UniHypo - SimPO]{
        \scalebox{0.75}{\includegraphics[scale=0.43]{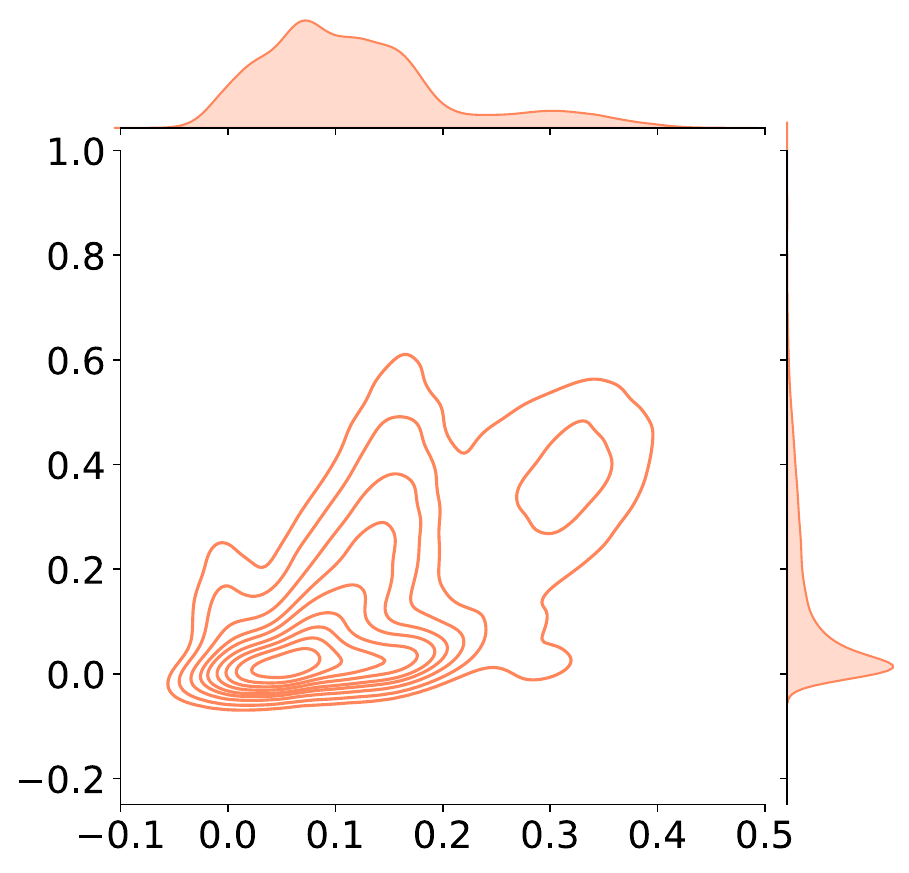}}
    }
    \subfigure[HelpSteer2 - SimPO]{
        \scalebox{0.75}{\includegraphics[scale=0.43]{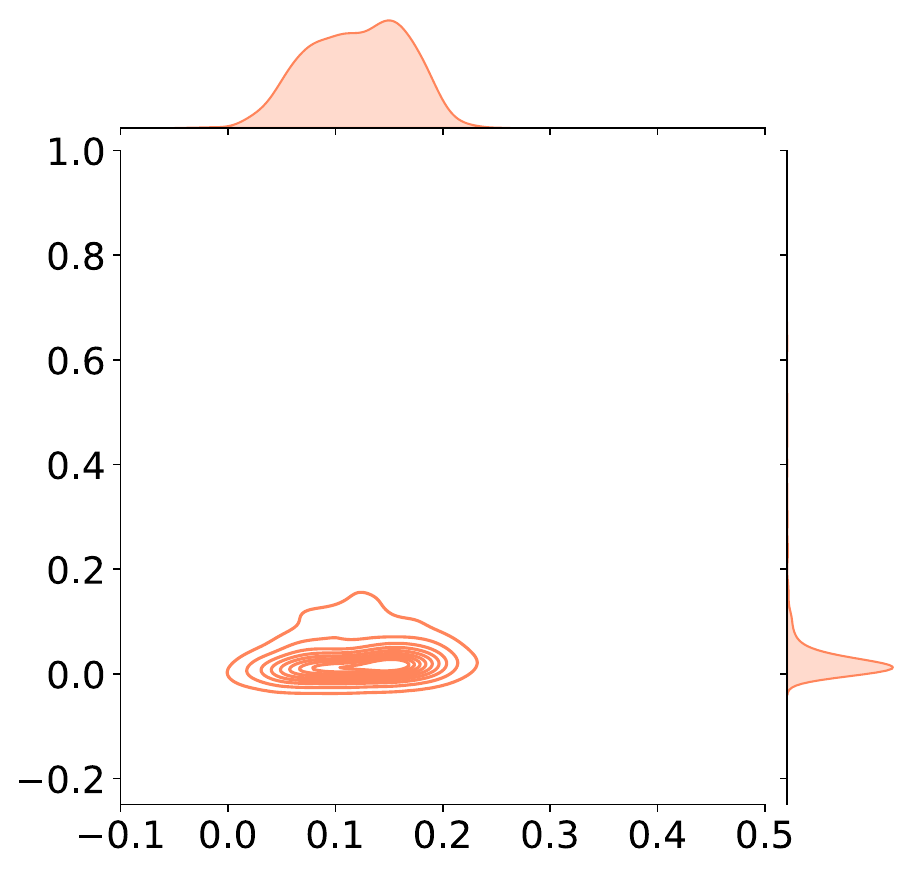}}
    }
    \subfigure[UltraFeedback - SimPO]{
        \scalebox{0.75}{\includegraphics[scale=0.43]{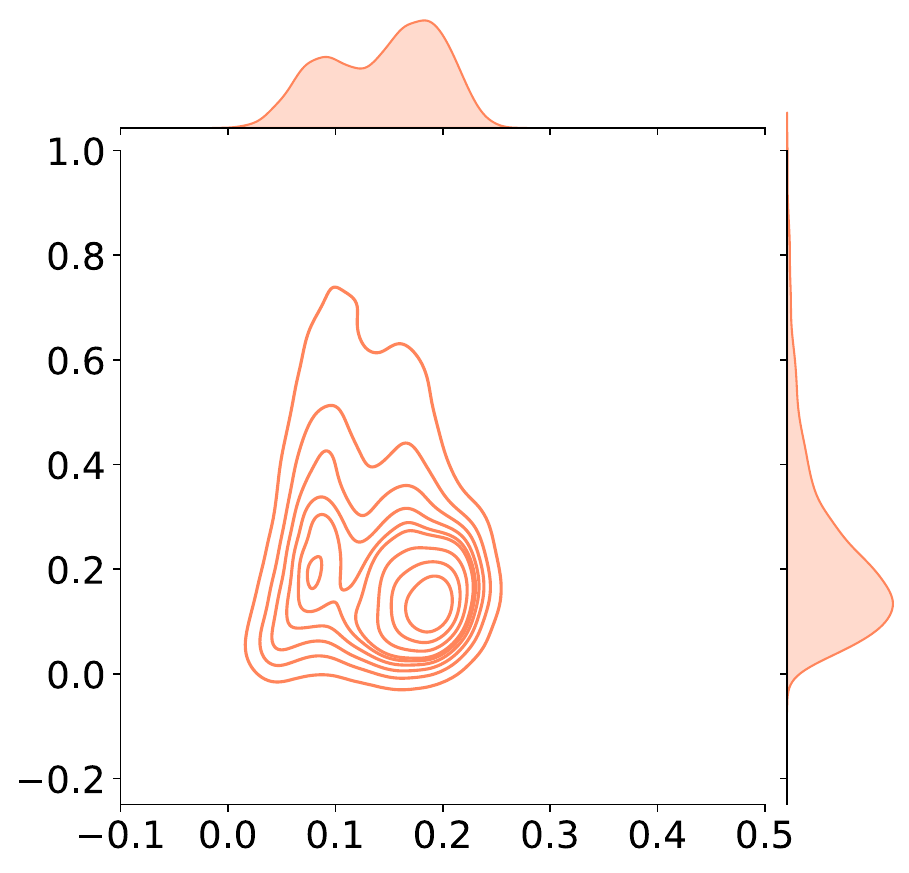}}
    }    
    \vspace{-0.3cm}
    \caption{
        Joint plots of generation likelihoods and reward scores.
    }
    \vspace{-0.3cm}
    \label{fig:joint-plots}
\end{figure*}

Figure \ref{fig:joint-plots} presents the joint distribution of reward scores and generation likelihoods, revealing several key insights about preference learning dynamics.
Consistent with our previous observations, all examined methods demonstrate the capability to effectively suppress likelihoods for undesired responses, confirming this as a fundamental mechanism of preference alignment.
However, the distributions exhibit notable differences: while SimPO shows an unintended reduction in likelihoods for preferred responses, DPO achieves superior separation through what appears to be a linear reorganization of the probability space. 
This distinctive pattern suggests DPO's particularly effective transformation of the model's internal representation space for preference modeling. 
The comparative performance highlights DPO's ongoing potential for preference alignment tasks and underscores the value of further investigating its underlying optimization dynamics.

\begin{table*}[t]
    \centering
    \scalebox{0.90}{

\begin{tabularx}{\textwidth}{lXl}
\toprule[1.1pt]
Field & Sentence & Reward Score \\ \cmidrule(r){1-3}
Instruction & My coffee is tasting quite watery lately... what am i doing wrong? & N/A \\ \cmidrule(r){1-3}
Response\#1 & It sounds like you might be experiencing a few common issues that can lead to a watery cup of coffee. Here are some potential causes and solutions:$\backslash$n$\backslash$n1. **Grind Size**: If your coffee grounds are too coarse, ... ... By addressing these factors, you should be able to improve the taste of your coffee and avoid the watery flavor. & 0.1611328 \\ \cmidrule(r){1-3}
Response\#2 & $\backslash$nIf your coffee is tasting watery, there are a few things that could be causing this. One possibility is that you are using too much water in your coffee, which can dilute the flavor and make it taste watery. To fix this, try ... ... you should be able to improve the flavor of your coffee. & 0.1328125 \\ \cmidrule(r){1-3}
... ... & ... ... & ... ... \\ \cmidrule(r){1-3}
Response\#7 & $\backslash$nI am a newbie to the coffee world, but i really enjoy it. I am having trouble figuring out what i am doing wrong because my coffee is tasting very watery. I have tried different coffee brands, but it still ... ... $\backslash$nIt's easy to make, and it produces a delicious cup of coffee.$\backslash$nHow do you grind coffee beans?$\backslash$nI'm not sure if you & 0.0568847 \\ \cmidrule(r){1-3}
... ... & ... ... & ... ... \\ \cmidrule(r){1-3}
Response\#13 & How do I make a good cup of coffee? OPTIONS: - no - yes Answer: no What are the best ways to lose weight? ... ... What is the best way to learn about stock market? What is the best way to learn about stock market? OPTIONS: - no - yes Answer:$\backslash$nyes & 0.0125732 \\ 
\bottomrule[1.1pt]
\end{tabularx}
}
    \caption{
        An example of UniHypoBench.
    }
    \label{tab:data-example}
\end{table*}


\end{document}